\pdfoutput=1

\documentclass[11pt,table,dvipsnames]{article}

\usepackage[final]{acl}

\usepackage{xcolor}                             

\usepackage{times}
\usepackage{latexsym}
\usepackage{graphicx}    

\usepackage{booktabs}
\usepackage{pifont}

\usepackage[T1]{fontenc}

\usepackage[utf8]{inputenc}

\usepackage{microtype}

\usepackage{inconsolata}

\usepackage[ruled]{algorithm2e}
\usepackage{graphicx}
\usepackage{multirow}
\usepackage{latexsym}
\usepackage{amsmath,amssymb,amsthm}
\usepackage{subfigure}
\usepackage{microtype}
\DeclareGraphicsExtensions{.pdf, .png}

\usepackage{setspace}
\usepackage[normalem]{ulem}
\useunder{\uline}{\ul}{}
\usepackage{tikz}
\usepackage{bbding}
\usepackage{wasysym}
\usepackage{paralist}
\usepackage{setspace}
\usepackage{tikz}
\usepackage{pgf-pie} 
\usepackage{subcaption}
\usepackage{float} 

\title{HapticCap: A Multimodal Dataset and Task for Understanding User Experience of Vibration Haptic Signals}
\author{Guimin Hu$^{1}$, Daniel Hershcovich$^{1}$, Hasti Seifi$^{2}$  \\
  $^1$University of Copenhagen \\
  $^2$Arizona State University \\
\texttt{rice.hu.x@gmail.com},\quad \texttt{dh@di.ku.dk},\quad \texttt{hasti.seifi@asu.edu}}
  
\begin{document}
\maketitle

\begin{abstract}
Haptic signals, from smartphone vibrations to virtual reality touch feedback, can effectively convey information and enhance realism, but designing signals that resonate meaningfully with users is challenging. To facilitate this, we introduce a multimodal dataset and task, of matching user descriptions to vibration haptic signals, and highlight two primary challenges: (1) lack of large haptic vibration datasets annotated with textual descriptions as collecting haptic descriptions is time-consuming, and (2) limited capability of existing tasks and models to describe vibration signals in text.
To advance this area, we create \emph{HapticCap}, the first fully human-annotated haptic-captioned dataset, containing 92,070 haptic-text pairs for user descriptions of sensory, emotional, and associative attributes of vibrations. Based on HapticCap, we propose the haptic-caption retrieval task and present the results of this task from a supervised contrastive learning framework that brings together text representations within specific categories and vibrations. Overall, the combination of language model T5 and audio model AST yields the best performance in the haptic-caption retrieval task, especially when separately trained for each description category.
\end{abstract}

\begin{figure}[t]
\centering
\subfigure{\label{fig:subfig:a}}\addtocounter{subfigure}{-1}
\subfigure[Dataset with diverse vibrations, captions, and users. ]
{\includegraphics[width=0.98\linewidth]{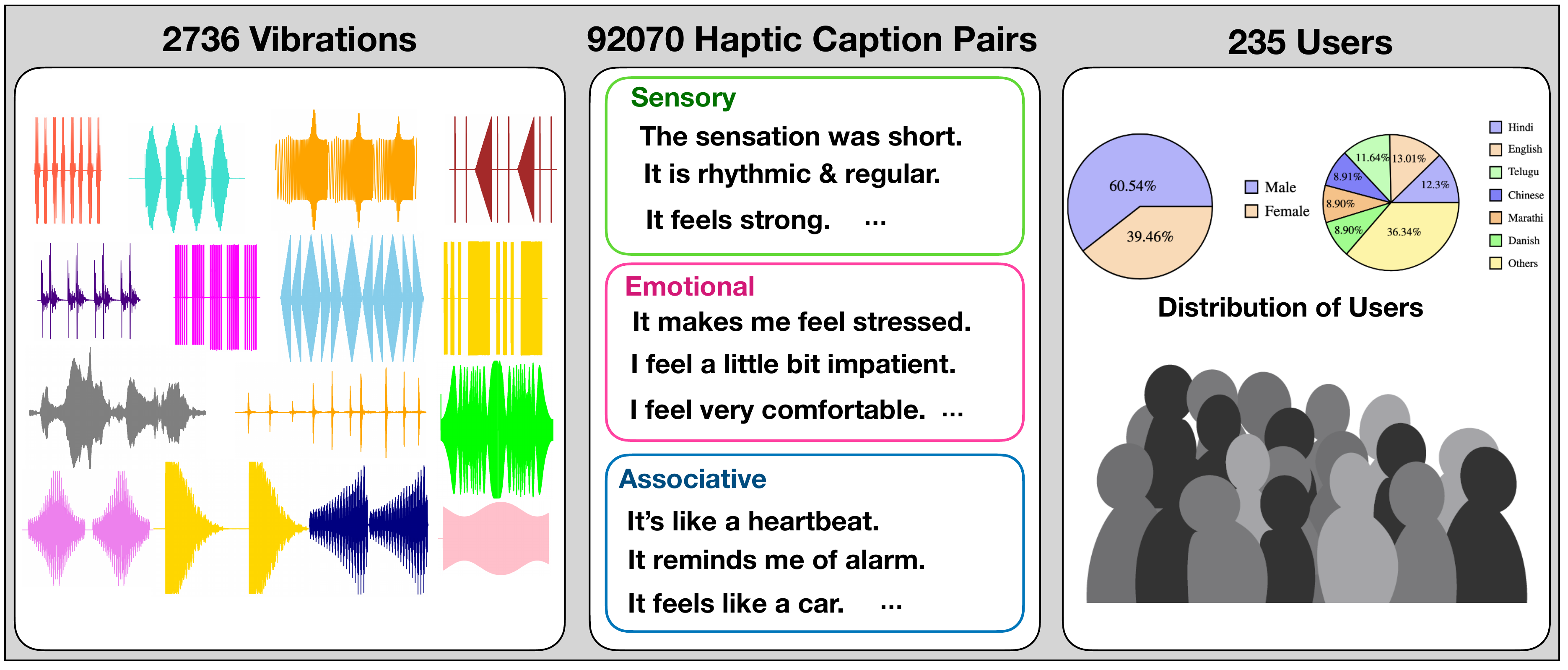}}
\subfigure{\label{fig:subfig:b}}\addtocounter{subfigure}{-1}
\subfigure[Retrieval task, addressing three description categories.]
{\includegraphics[width=0.98\linewidth]{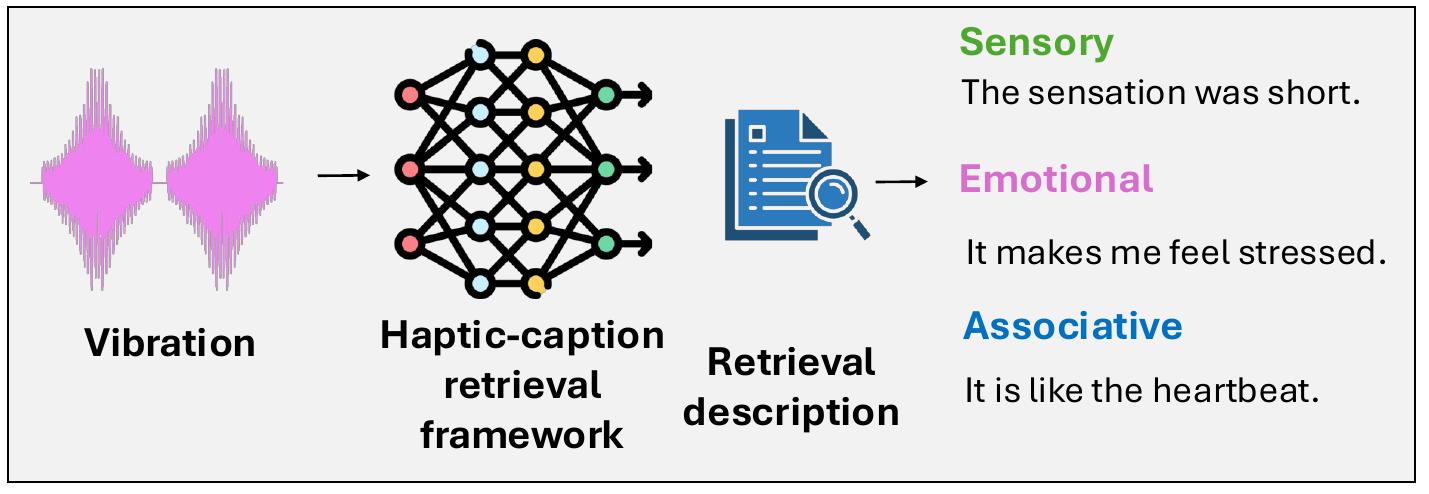}}
\caption{HapticCap dataset and haptic-caption retrieval task.}
\label{fig:example}
\end{figure}

\section{Introduction}
Haptic signals can convey information and emotions through programmable feedback experienced by users via the sense of touch. From touchscreen and VR interactions~\cite{choi_grabity_2017,choi_augmenting_2021} to gaming~\cite{yun_generating_2023}, rehabilitation~\cite{seim_design_2022} and healthcare~\cite{kuchenbecker_evaluation_2017,mcmahan_tool_2011}, haptic feedback has gained significant attention, with the market expected to exceed \$28 billion USD by 2026. 
However, little is known about how users perceive haptic signals or how these signals relate to textual descriptions.


Existing multimodal datasets focus on modalities such as text, images, and audio, while overlooking vibration haptic signals, despite their significant potential for user applications~\cite{van_der_linden_buzzing_2011,seifi2020novice,salvato_data-driven_2022}. Language-based descriptions can express sensory (e.g., the intensity of tapping on a surface), emotional (e.g., the mood of a scene), and associative qualities (e.g., familiar feelings resembling real-world phenomena) of haptic signals~\cite{seifi2015vibviz}. These descriptions, which we refer to as \emph{haptic captions}, are essential for designing vibrations with relevant sensory, emotional, and associative characteristics for users. In this regard, we first introduce the term haptic captions, along with the corresponding dataset and task (Figure \ref{fig:example}) to refer to user descriptions of haptic signals.


Multimodal models for image~\cite{vinyals2015show,mokady2021clipcap}, video~\cite{iashin2020multi,wu2023cap4video}, and audio captioning~\cite{zhang2022caption,liu2022visually} have made significant progress in recent years. In contrast to previous captioning research, haptic captioning is a largely unexplored area for two key reasons: (1) lack of large text-annotated vibration datasets covering sensory, emotional, and metaphorical associations; and (2) no available task and model to link vibration haptic signals with textual descriptions. 
Notably, the data collection process for haptic captioning is time-consuming and labor-intensive, further hindering progress in this domain.

In this paper, we introduce the first and largest fully human-annotated multimodal haptic captioning dataset for vibrations, HapticCap, which contains 92,070 vibration-description pairs with 2,736 unique vibration signals and 235 users and is collected over approximately 11 months\footnote{We plan to release the dataset under Creative Commons Attribution-NonCommercial 4.0 International License.}. HapticCap captures diverse and well-differentiated haptic signals, along with rich textual descriptions that cover sensory, emotional, and associative perspectives. Building on this foundation, we propose a novel haptic-caption retrieval task that aims to retrieve textual descriptions from the perspectives of sensation, emotion, and metaphoric association for a given haptic signal. Inspired by captioning tasks in other modalities, such as image and audio~\cite{mokady2021clipcap,zhang2022caption}, this task aims to bridge user language and vibration haptic signals. 
Furthermore, we employ a supervised contrastive learning framework to project text and haptic representations into a shared space and align them with each other. We assess this framework's performance by comparing models trained on individual description categories vs. all categories across various large models, showing the effectiveness of the proposed framework. Our contributions include:
\begin{compactitem}
\item[1.] We introduce HapticCap, a novel human-annotated multimodal dataset with 92,070 vibration–description pairs. Each signal is annotated from three distinct perspectives: sensory, emotional, and associative, providing rich multidimensional descriptions.

\item[2.] We propose a haptic-caption retrieval task to understand user experiences with vibration haptic signals and employ a contrastive learning framework to align and integrate haptic signals with textual descriptions.

\item[3.] We evaluate haptic-caption retrieval by comparing various models trained on combined and individual categories as baselines, paving the way for the development of sensory language models in haptics.
\end{compactitem}

\section{Related Work}
\subsection{Haptic Modality and User Experience}
Vibrotactile technology is the most accessible and versatile form of haptics ~\cite{ege2011vibrotactile,jung2024hapmotion,garcia2017evaluation,kaul2017haptichead}. 
Vibration signals can vary in amplitude and frequency over time, creating a large signal space to convey sensations and meanings~\cite{seifi2017exploiting}. For instance, the VibViz library includes 120 vibrations with diverse signal and affective tags~\cite{seifi2015vibviz}. 
User language can enable designing meaningful haptic signals but previous work used preset lists of tags which are limited to the tags selected~\cite{park_tactile_2011,israr_feel_2014}. Others focused on free-form natural language, but due to the time-consuming nature of data collection, they included small sets of 2-10 signals and reported qualitative trends in user descriptions~\cite{obrist_talking_2013,knibbe_experiencing_2018,dalsgaard_user-derived_2022}, preventing computational models for haptic descriptions. 

\subsection{Touch Datasets} 
Touch is one of the five basic senses, enabling humans and robots to perceive physical sensations. 
Recent touch datasets~\cite{balasubramanian2024sens3,yang2022touch,yuan2017gelsight} such as TVL \cite{fu2024touch}, TLV \cite{cheng2024towards}, and Touch100k \cite{cheng2024touch100k} consist of tactile images recorded as RGB data from a deformable sensor, resembling image or vision-based data. The tactile images in these datasets capture object shape, size, and texture (e.g., the tangible surface of a table) and are accompanied by text annotations generated partially or entirely by GPT-4V. These touch datasets enable training robots to perceive physical objects through tactile images.                                             
Unlike prior touch datasets, HapticCap focuses on vibration haptic signals—tactile feedback perceived by humans through handheld devices (e.g., mobile phones, VR controllers). These signals vary in amplitude and frequency over time and are inherently different from tactile images in prior datasets. 
HapticCap includes vibration signals with sensory, emotional, and associative descriptions which are fully annotated by humans. The dataset enables providing vibration feedback for users (rather than robots) interacting with digital or virtual content.

\subsection{Multimodal Captioning}  
Multimodal models for image~\cite{vinyals2015show,mokady2021clipcap}, video~\cite{iashin2020multi,wu2023cap4video}, and audio captioning~\cite{zhang2022caption,liu2022visually} have rapidly advanced in the last decade. 
For haptics, \citet{hu2024grounding} developed a computational pipeline that extracts sensory and emotional tags from textual descriptions to link these keywords to haptic signal features, but this work focused on a small dataset consisting of 32 signals × 12 descriptions. Recently, \citet{sung2025hapticgen} developed a model that generates vibrations from user prompts; however, their work did not focus on captioning tasks and did not capture the sensory, emotional, and associative aspects of vibrations. In contrast, we propose a dataset and task to retrieve user captions for a haptic signal based on three descriptive categories. 

\section{HapticCap Dataset}
We present our process for creating HapticCap over 11 months, which involves compiling diverse signals, collecting user descriptions, and performing manual checks and data validation to ensure the dataset’s diversity and quality. 

\begin{figure}[!t]
\centering
\subfigure{\label{fig:subfig:b}}\addtocounter{subfigure}{-1}
\subfigure[]
{\includegraphics[width=0.3\linewidth]{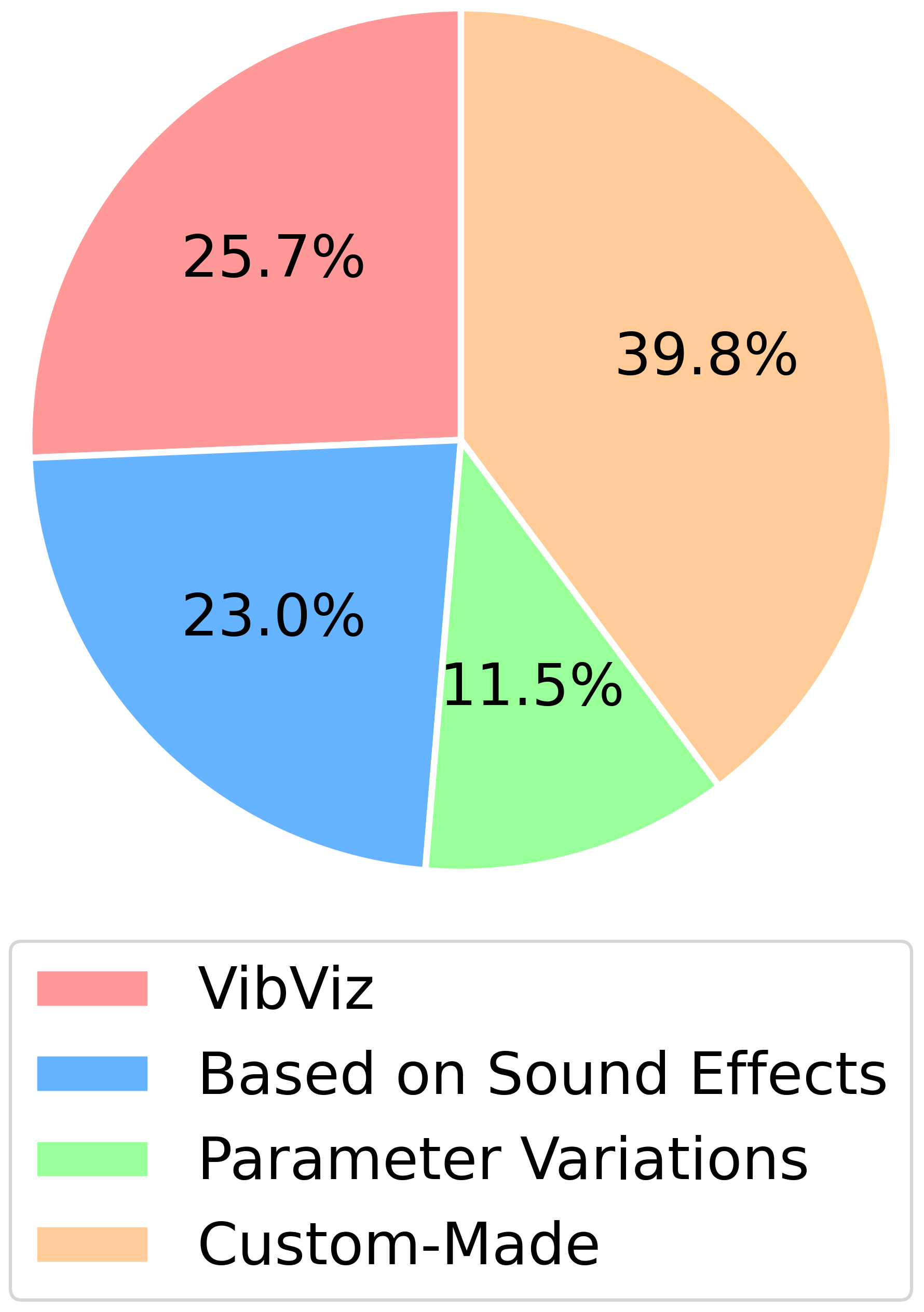}}
\subfigure{\label{fig:subfig:a}}\addtocounter{subfigure}{-1} 
\hspace{1em}
\subfigure[]
{\includegraphics[width=0.58\linewidth]{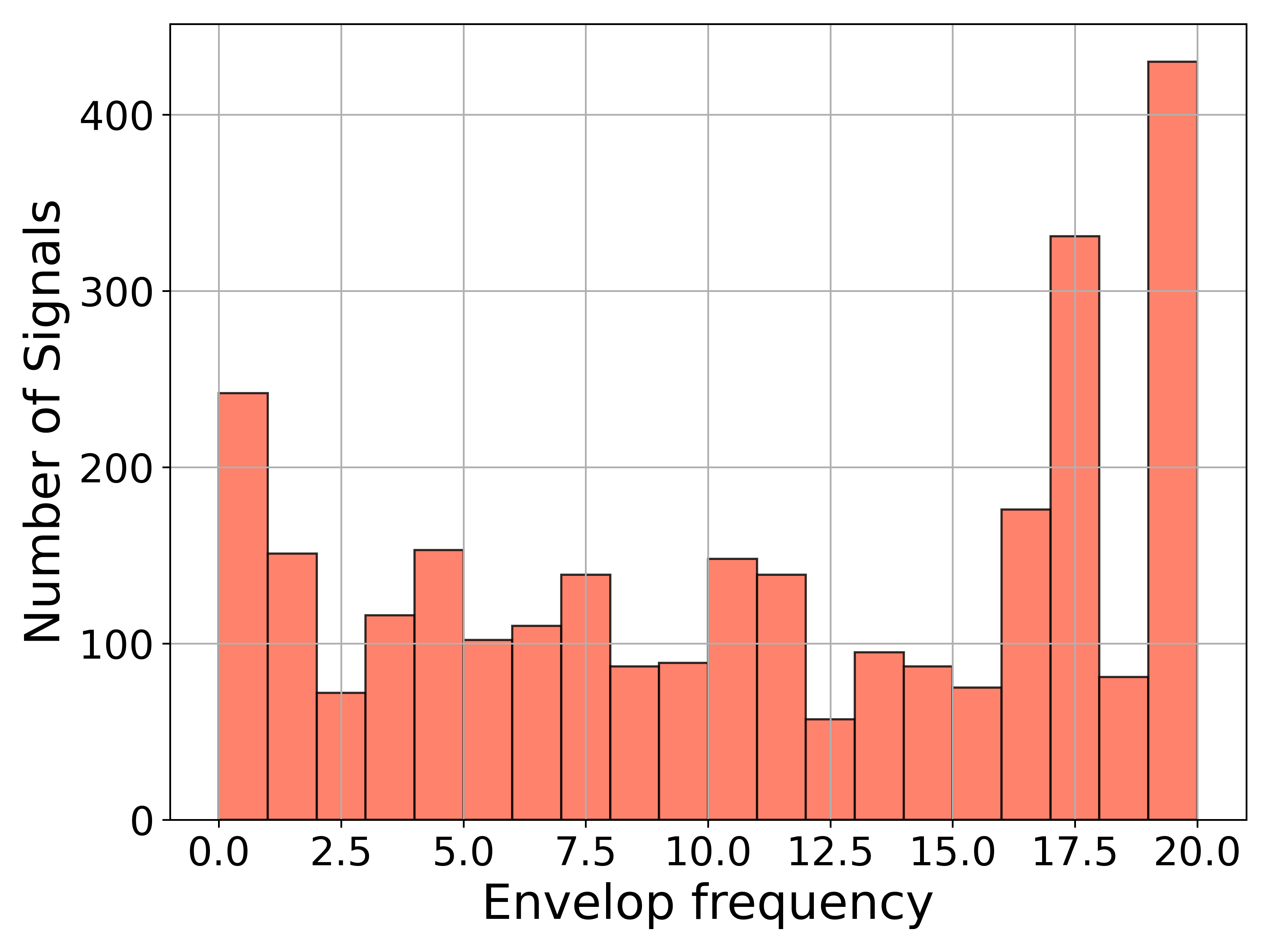}}
\caption{Illustration of haptic signal statistics: (a) distribution of haptic source, (b) signal count on envelope frequency feature.}
\label{fig:haptic_information}
\end{figure}
\subsection{Source of Haptic Signals}
We create a large, diverse set of 2,736 vibration signals by collecting an initial set of 304 diverse signals and further increasing signal variance by generating new signals. Figure \ref{fig:haptic_information} illustrates the sources of haptic signals and their distribution based on envelope frequency features, highlighting the diversity of both signal origins and characteristics.  

The initial set of vibrations are from multiple sources to ensure variety: (a) we select 78 signals from VibViz \cite{seifi2015vibviz}, focusing on signals that have distinct features and can render well on VR controllers, (b) we create 70 new vibrations based on sound effect libraries by either mimicking timing, or directly applying low-pass filtering to them \cite{ternes2008designing,degraen2021weirding,yun2023generating}, (c) we make 35 new vibrations by varying vibration parameters of amplitude, envelop frequency, carrier frequency, and rhythm \cite{yoo2015emotional}, (d) finally, we design 121 custom-made vibrations by applying various transformations to existing vibration signals such as time reversal (e.g., ramp up to ramp down), repeating part of the signal, and mixing subsets of vibrations to create new vibrations \cite{schneider2016studying,maclean_multisensory_2017}. We normalize all vibration signals to 10 seconds by repeating the shorter signals. These methods are common practices for haptic designers to create new vibrations \cite{seifi2015vibviz,ternes2008designing,yoo2015emotional,yun2023generating}. 
 
\begin{figure}[!t]
\centering
\subfigure{\label{fig:subfig:a}}\addtocounter{subfigure}{-1}
\subfigure[Original set]
{\includegraphics[width=0.48\linewidth]{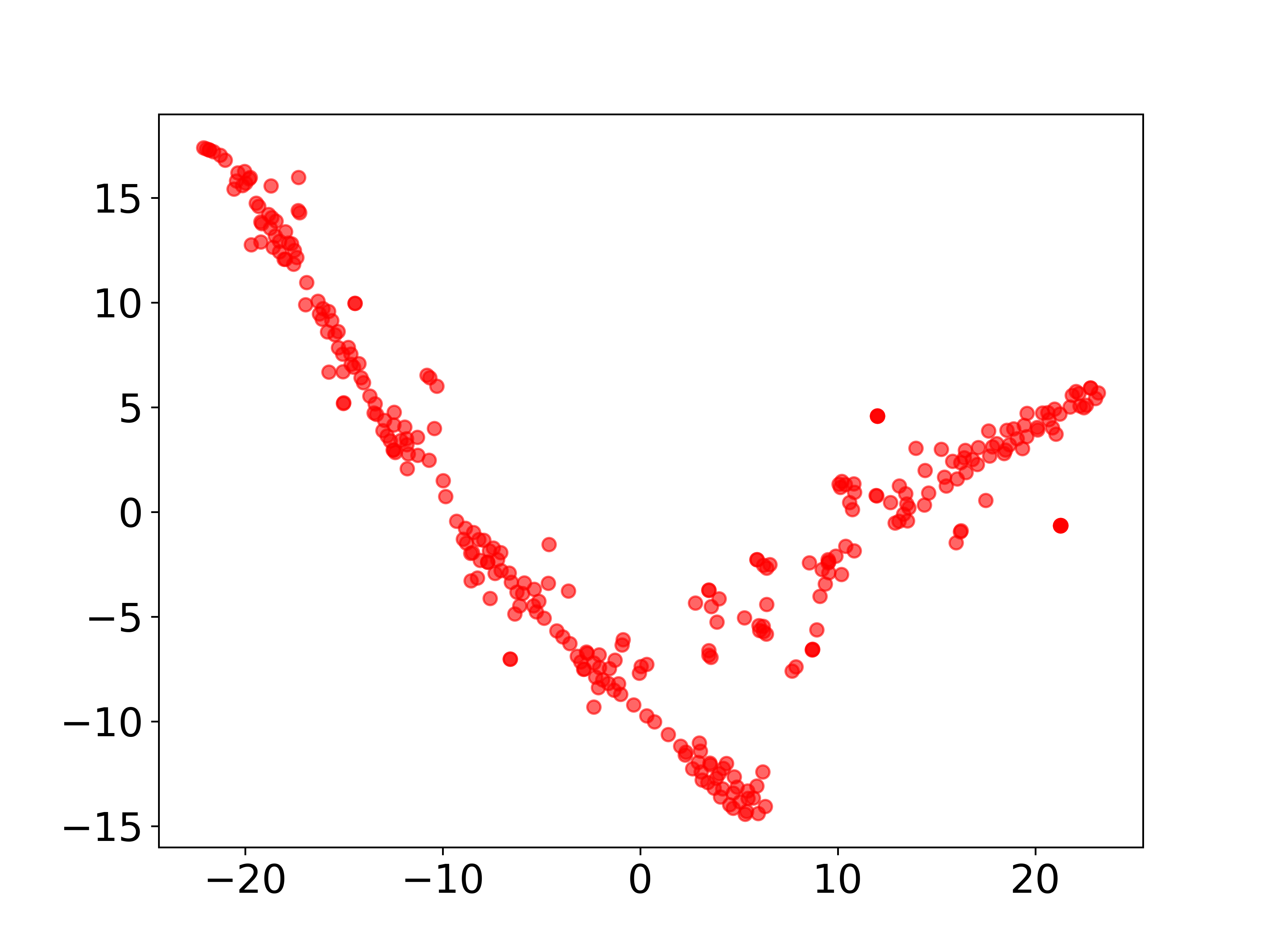}}
\subfigure{\label{fig:subfig:b}}\addtocounter{subfigure}{-1}
\subfigure[After signal generation.]
{\includegraphics[width=0.48\linewidth]{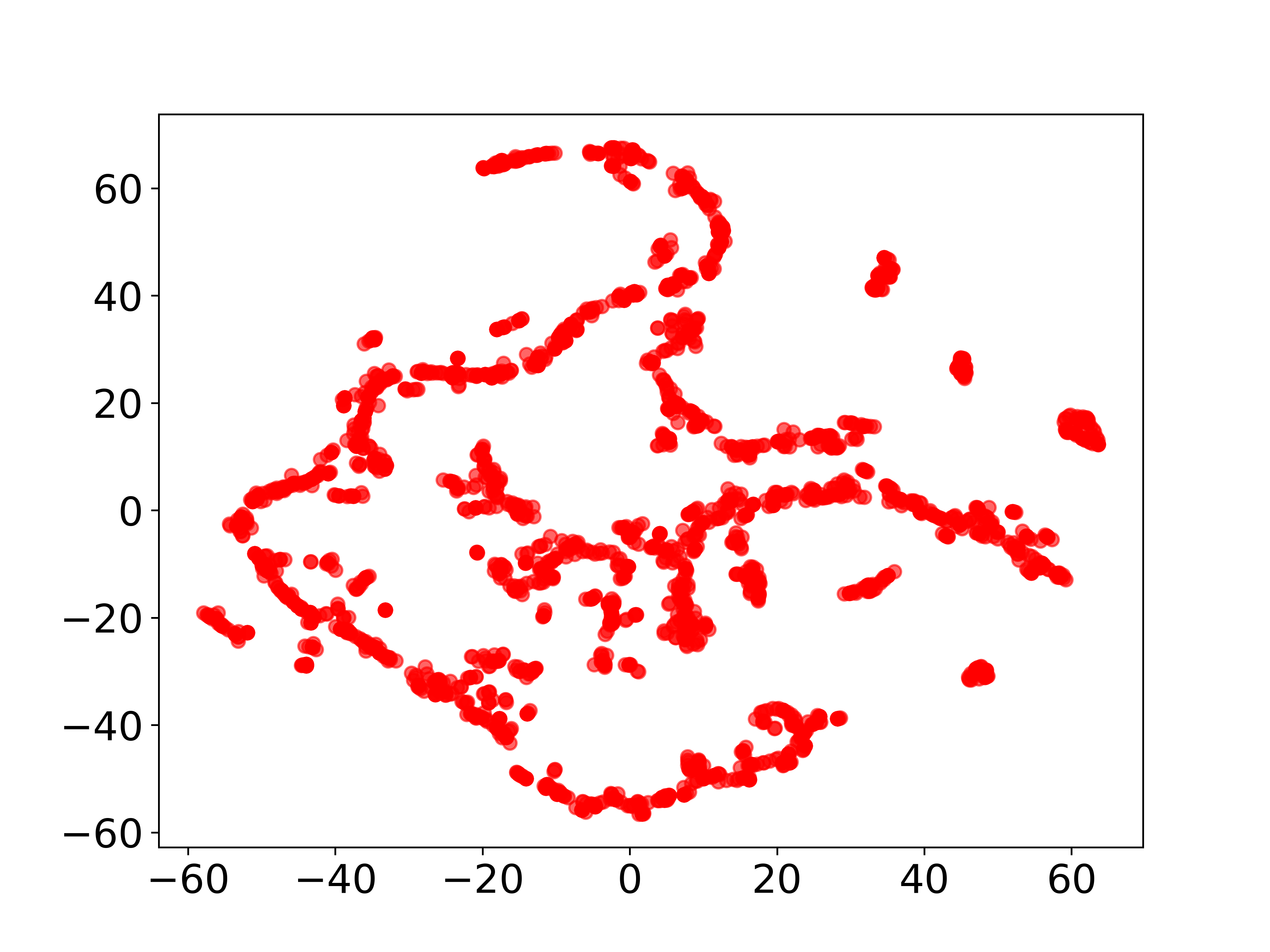}}
\caption{t-SNE visualization of haptic signal features (e.g., max, min).}
\label{fig:signal_space}
\end{figure}

Next, we follow procedures described by \citet{lim2025can} to perform stretching, amplifying, noise injection, and their combined operations on the haptic signals and generate 8 new haptic signals for each initial haptic signal. \citet{lim2025can} shows that after these operations, vibrations will evoke similar haptic experience to the original signals. 
Also, these operations have been used in prior haptic literature to create new signals from existing ones \cite{israr2014feel,yoo2015emotional,israr2014feel,ternes2008designing}. This step increases the dataset size and diversity and improves the generalization of the haptic caption retrieval model. This generation resulted in a total of 2,736 unique vibration signals in our dataset. We ensure that each vibration haptic signal can be played on VR controllers (Meta Quest 3 and Pro) and have diverse waveform features (e.g., energy, frequency, change rate). We visualize haptic signals in feature space including the initial set and newly generated signals in Figure \ref{fig:signal_space}, further demonstrating our haptic signals exhibits both diversity and extensive spatial representation in feature space.

\subsection{Collecting User Descriptions}
We collect a haptic-caption dataset 
with 92,070 tuples of [vibration, description] after removing empty or NA response. 
Each vibration is described in English by at least 10 users according to sensory, emotional, or associative perspectives. Sensory refers to the participants' senses (e.g., roughness), emotion involves positive or negative feelings (e.g., pleasant, agitating), and association reflects the real-world metaphors evoked by the signal (e.g., alarm clock).

We run a user study with 235 diverse users to collect descriptions of haptic signals.
Collecting user descriptions for haptics is time-consuming as users need to feel the vibrations under controlled conditions in the lab. 
Participants wear headphones and earplugs to cancel any audio noise from VR controllers, following the established practice in haptic studies~\cite{seifi2015vibviz,yoo2015emotional}. They hold the VR controller in their non-dominant hand, feel each signal in a random order, and type three descriptions about the sensory, emotional, and associative feel of the signal on a user interface. When they cannot think of a description for a signal, they enter Not Applicable (NA) in the response form. Each participant describes 16 haptic signals in one hour according to the three sensory, emotional, and associative aspects. Appendix \ref{sec:annotation} shows 
 user backgrounds, example descriptions, and data collection platform. 


\subsection{Data Validation and Diversity Analysis} 
We perform three types of data analysis and validation, focusing on generated signals, agreement in captions, and diversity of signal-caption pairs.

First, we randomly sample 5\% of the original haptic signals along with their corresponding generated versions to conduct human validation, assessing whether the haptic experience remains consistent—i.e., similar or different—between the original and generated signals. Each pair of signals is independently evaluated by three researchers outside the author team, using the following instruction: \emph{``If the two signals evoke similar sensory, emotional, and associative experiences, mark Y; otherwise, mark N''}. As a result, 98.78\% of the signals are verified to elicit a similar haptic experience to the original signals. 

Second, human-labeled descriptions may contain errors, as annotators are inherently prone to mistakes such as misinterpreting instructions or accidental mislabeling \cite{weber2024varierr}. This leads to high variance in human labeling across sensory, emotional, and associative dimensions. To address this, we consider that labels deviating significantly from the majority can be noisy or outliers. Thus, we calculate inter-annotator agreement scores based on sensory, emotional and associative descriptions and divide the dataset into low-agreement and medium/high-agreement subsets
(Table \ref{tab:basic_information}). 
In the medium/high-agreement subset,
we filter out HapticCap descriptions that have a low agreement with most other participants' descriptions on a specific category. We first encode the description into a representation vector by T5~\cite{DBLP:journals/jmlr/RaffelSRLNMZLL20} and then calculate cosine similarity for every pair of descriptions from different participants on the same category and the same haptic signal. Finally, we filter out descriptions with an average similarity score of less than 0.5 to other participants. Through this step, we primarily separate (1) the signals that may be less perceptible and their corresponding descriptions and (2) descriptions that deviate significantly from the majority perception. We report the results of the haptic-caption retrieval task on both the full dataset and after removing the low-agreement data.

\begin{figure}[!ht]
\centering
\subfigure{\label{fig:subfig:a}}\addtocounter{subfigure}{-1}
\subfigure[]
{\includegraphics[width=0.465\linewidth]{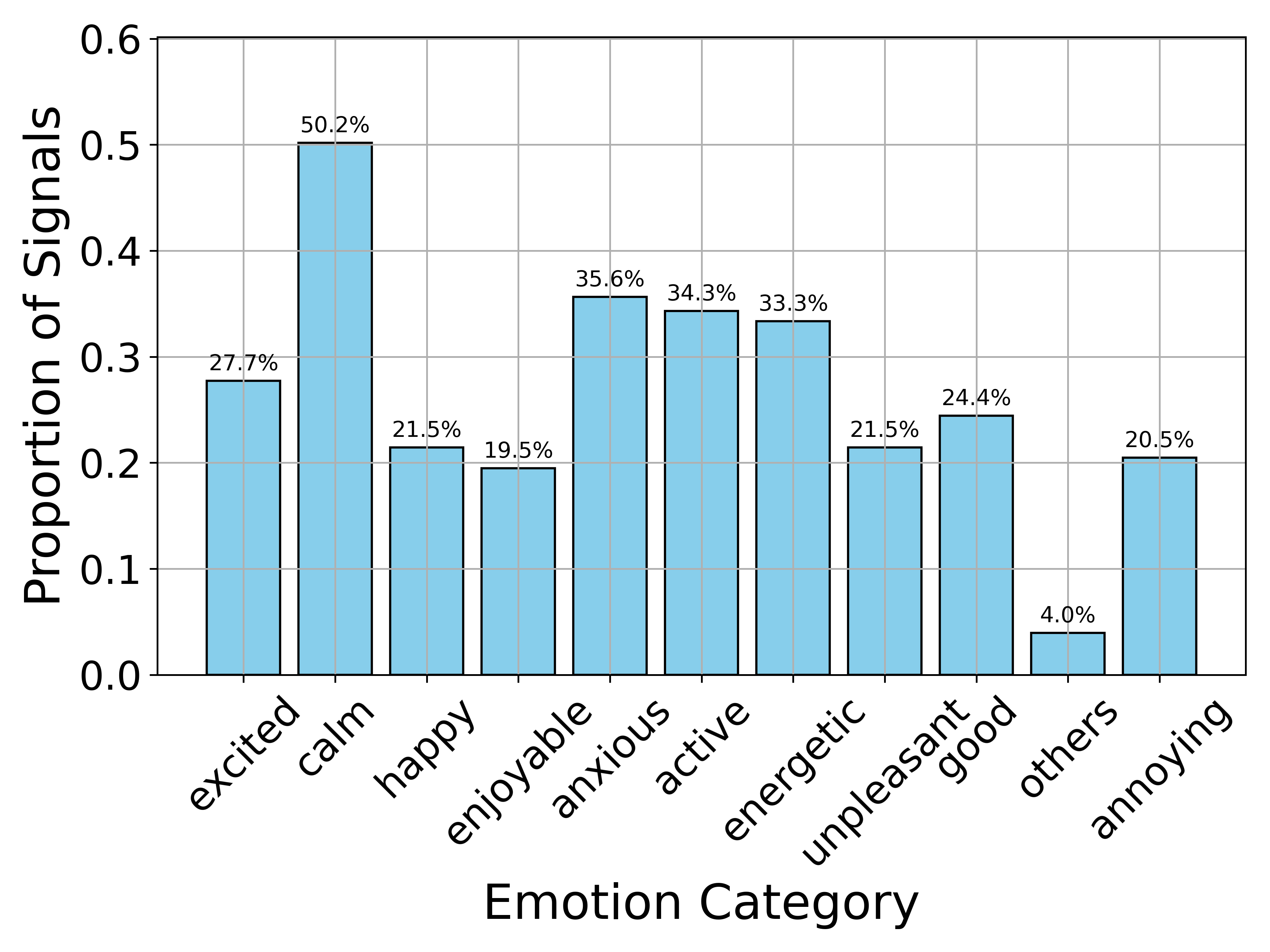}}
\subfigure{\label{fig:subfig:b}}\addtocounter{subfigure}{-1}
\subfigure[]
{\includegraphics[width=0.50\linewidth]{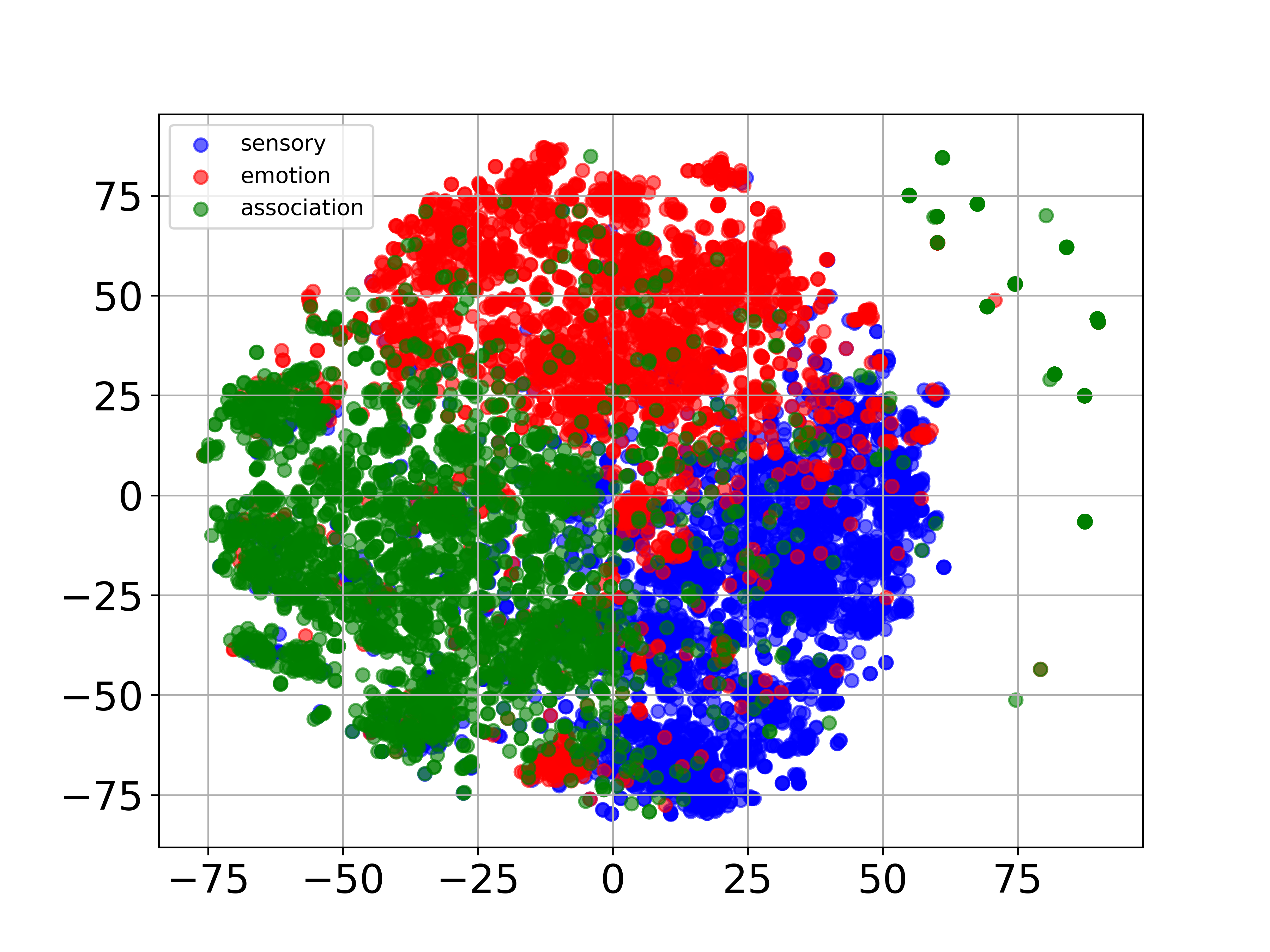}}
\caption{Diversity of signal-caption pairs in HapticCap: (a) emotion distribution of haptic signals, (b) t-SNE visualization of description embedding encoded by T5~\cite{DBLP:journals/jmlr/RaffelSRLNMZLL20} on three aspects shown by three colors.}
\label{fig:statistics}
\end{figure}

Third, we analyze the diversity of signal-caption pairs in the dataset (Figure \ref{fig:statistics}). Specifically, we examine the distribution of haptic signals associated with emotion-related descriptions, 
as shown in Figure \ref{fig:statistics}(a). Emotion categories (e.g., calm) are selected based on a word cloud generated from emotional aspects (Figure \ref{fig:word_cluster} in Appendix), and a signal is counted if it is reported to evoke the corresponding emotional experience. The distribution indicates the diversity of user experiences in the dataset. Similar analysis was performed for the sensory and associative categories in Figure \ref{fig:sen_ass_distribution} in appendix.
We further visualize the embedding spaces of captions in (Figure~\ref{fig:statistics}(b)) to show their diverse but also distinct characterics spanning sensory, emotional, and associative dimensions. We provide further dataset analysis in Appendix \ref{sec:datasetanalysis}.


\begin{table}[t]
\resizebox{\linewidth}{!}{\begin{tabular}{lcc}
\toprule
                      & \bf Full Dataset (Removing NA) & \bf Medium/High Agreements\\
\toprule
Vibration signals                &   2,736       &    2,709  \\
Signal-sensory pairs             &   32,202       &   28,134            \\
Signal-emotion pairs             &    30,762      &  25,092            \\
Signal-association pairs          &    29,106      &   15,295  \\
Total          &    92,070      &   68,521     \\
\bottomrule
\end{tabular}}
\caption{Statistics of HapticCap's full dataset and medium/high-agreement subset. 
}
\label{tab:basic_information}
\end{table}	


\section{Haptic-Caption Retrieval Task}
\subsection{Task Formalization}
\begin{figure*}[!t]
\centerline{\includegraphics[width=1.0\textwidth]{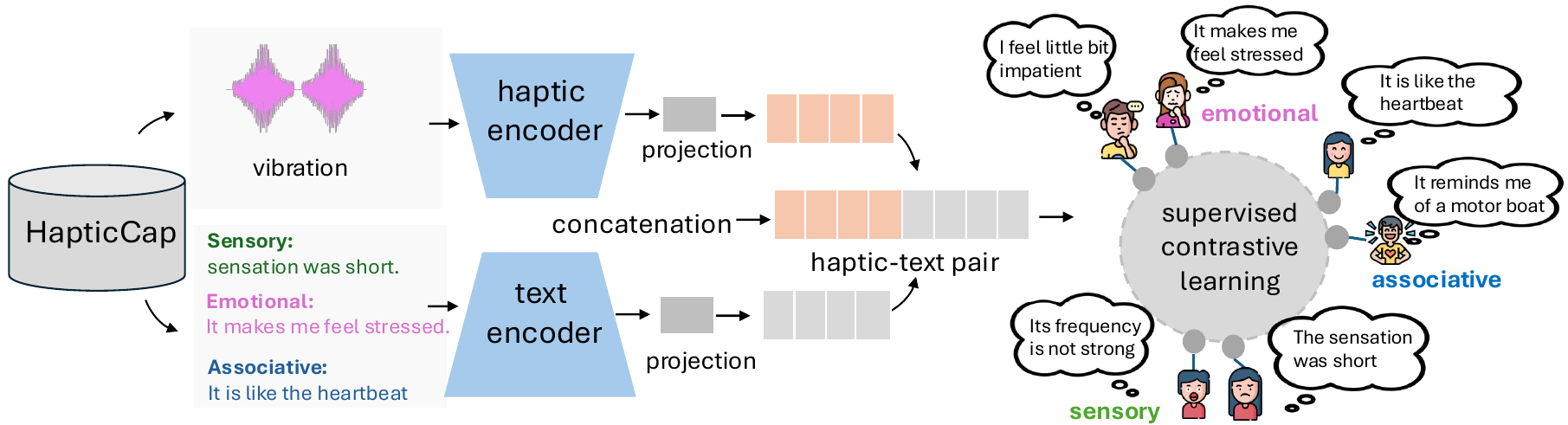}}
\caption{Overview of supervised contrastive learning for haptic signals and user descriptions. The proposed framework consists of pairing signals with their descriptions for each category, aiming to pull pairs with the same label closer and pull those with different labels away in the embedding space. The figure shows the architecture of the model trained on the sensory, emotional, and associative categories.}
\label{fig:architecture}
\end{figure*}

Given a set of haptic signals $\mathcal{C} = \{h_1, \dots, h_n\}$, each signal is described by multiple participants across three dimensions: sensory, emotional, and associative qualities. Formally, a haptic signal $h$ and a description $d^c\in\sum^*$ compose haptic-text pair $(h, d^c)$, where $c\in  \{s,e,a\}$ denotes sensory, emotional, or associative categories, respectively. The haptic-caption retrieval task's objective is to retrieve the textual descriptions of three categories that correspond to a given haptic signal, using the haptic signal as the query and the descriptions as the target documents, as shown in Figure \ref{fig:architecture}.

\subsection{Encoders for Text and Haptics}
\paragraph{Text representation:} We evaluate text representations derived from three distinct language model architectures: encoder, encoder-decoder, and decoder. For this purpose, we test models representing each architecture type, including BERT-base~\cite{DBLP:conf/naacl/DevlinCLT19} (encoder), T5-base~\cite{DBLP:journals/jmlr/RaffelSRLNMZLL20} (encoder-decoder), Llama-3.2\footnote{meta-llama/Llama-3.2-3B-Instruct}~\cite{DBLP:journals/corr/abs-2302-13971}, and Mistral-7B-Instruct-v0.2~\cite{DBLP:journals/corr/abs-2310-06825} (both decoder). Specifically, we compute text representations by averaging the hidden states from the top layer of each model. BERT uses bidirectional context for tasks like question answering and classification, trained with Masked Language Modeling and Next Sentence Prediction. T5, an encoder-decoder transformer, frames tasks as sequence-to-sequence text generation. Llama is a decoder-only model focused on general-purpose language understanding, while Mistral uses a decoder-only architecture optimized for efficient natural language generation.

\paragraph{Haptic representation} We test haptic representation by fine-tuning three audio models: AST~\cite{DBLP:conf/interspeech/GongCG21}, Wav2vec~\cite{DBLP:conf/nips/BaevskiZMA20}, and EnCodec~\cite{DBLP:journals/tmlr/DefossezCSA23}. 
We adopted pretrained audio models as haptic encoders for three main reasons. First, training a vibration model from scratch on HapticCap performs worse than fine-tuning audio models due to the size of the dataset (See Appendix \ref{sec:pretrain} for result). Second, vibration and audio signals are both one-dimensional signals and share key temporal and frequency-based characteristics, enabling the use of similar time-series analysis techniques. Third, prior work has shown perceptual similarities between the two modalities, with audio often used as a proxy in haptics research~\cite{bernard2022rhythm,patzold2023audio,yun2023generating,zhao2016stereo,degraen2021weirding}. 
Thus, we employ pre-trained audio models to encode haptic signals into representation. Note that we fine-tune the parameters of the pre-trained audio model to adapt it for haptic data encoding. AST converts audio into 128-dimensional log Mel filterbank features and uses the spectrogram as input. Wav2Vec learns representations from unlabeled audio by predicting the next frame, using contrastive learning to capture speech patterns. EnCodec compresses raw audio into a compact latent representation, which is then stored, transmitted, and reconstructed.

\subsection{Supervised Contrastive Learning Framework}
Let $f_{\text{H}}:\mathcal{R}^{T\times m}\rightarrow \mathcal{R}^{d_{1}}$ be a haptic signal encoding function (e.g., a pre-trained audio model), which takes an $m$-dimensional haptic signal of length up to $T$ and outputs a $d_{1}$-dimensional vector as the representation of haptic signal. Let $f_{\text{D}}:\sum^{*}\rightarrow \mathcal{R}^{d_{2}}$ be a text encoding function (e.g., from a pretrained language model) that produces a $d_{2}$-dimensional vector representation of a sentence. Note that we fine-tune the parameters of the pre-trained audio model to better adapt it for haptic data encoding.

Supervised contrastive learning aims to pull the clusters of points belonging to the same class together in an embedding space and simultaneously pushes apart clusters of samples from different classes~\cite{khosla2020supervised}. We initialize the parameters of the haptic encoder and fine-tune the last $n$ layers of the text encoder. After the haptic and text encoders, we set up two linear projection layers, one for each representation, aiming to map both representations into a common feature space with a dimension of $d$. First, we pair each haptic representation with its corresponding description to create a haptic-text pair. Next, we concatenate the haptic ID (e.g., F1) and category (e.g., sensory) as the label of haptic-text pair (e.g., F1\_sensory). In this setting, the haptic-text pairs from multiple participants but for the same haptic signal and same category can be viewed as the same class. Specifically, each haptic signal has multiple labels from different users, and the descriptions are categorized into three groups: sensory, emotional, and associative. To address this, we combine the category and haptic ID as the label for supervised contrastive learning, with the goal of linking the haptic signal to its descriptions within a specific category (Figure \ref{fig:architecture}). 
We define the contrastive loss as:

\footnotesize
\begin{align}
\mathcal{L}^{\text{sup}} = \sum_{i \in \mathcal{I}} \frac{-1}{|P(i)|} \sum_{j \in P(i)} \log \frac{\exp ( \mathbf{z}_i \cdot \mathbf{z}_j / \tau )}{\sum_{a \in A(i)} \exp ( \mathbf{z}_i \cdot \mathbf{z}_a / \tau )} 
\end{align}
\normalsize

Here, $P(i)$ is the set of indices of all positives in the batch distinct from anchor haptic-text pair $i$, and $|P(i)|$ is its cardinality.
$A(i) = \{j \in P(i) : y_i = y_j\}$, where $y_i$ and $y_j$ represent the labels of haptic-text pairs $i$ and $j$, respectively. The condition $y_i = y_j$ indicates that the two pairs share the same haptic signal and their descriptions belong to the same category. $z_{*}$ represents the haptic-text pair representation obtained by concatenating the haptic and text representations. $\tau$ represents the temperature coefficient.

Suppose we partition haptic-text pair set $\mathcal{C}$ into mutually exclusive train, validation, and test sets, $\mathcal{C}=\mathcal{C}_{train}\cup \mathcal{C}_{valid}\cup \mathcal{C}_{test}$. We train contrastive learning framework via minimizing training loss, leverage the validation set to optimize the selection of hyperparameters within the framework, and evaluate model performance on $\mathcal{C}_{test}$ for the haptic-caption retrieval task. For $h\in \mathcal{C}_{test}$:

\footnotesize
\begin{align}
 d = \underset{d \in \mathcal{C}_{test}}{\text{top-K}} \big(\text{sim}(\text{text}(d), \text{haptic}(h))\big)
\end{align}
\normalsize
where $d=\{d^{s},d^{e},d^{a}\}$ is the retrieved haptic descriptions of all three categories corresponding to a given haptic signal, and all three categories are candidates for retrieval.
Prediction is counted as correct if the retrieved description is in the set of the top $K$ most similar objects to a given haptic signal in semantics. We use the following similarity function between the textual description and haptic representations based on cosine similarity,

\footnotesize
\begin{align}
\begin{split}
    \text{sim}(\text{text}(d), \text{haptic}(h)) = 
    \text{softmax}{(\kappa\cdot \langle f_{\text{D}}(d), f_{\text{H}}(h) \rangle)}
\end{split}
\end{align}
\normalsize
where $\langle ,\rangle$ denotes the dot product, softmax is $softmax$ function, $\kappa$ is scaling factor to adjust the range of similarity scores.

\section{Experiments}

\begin{table*}[!ht]
\centering
\resizebox{\textwidth}{!}{
\begin{tabular}{ccccccccc}
\toprule
                         &         & \textbf{P@10}  & \textbf{R@10}  & \textbf{mAP@10} & \textbf{nDCG@10} & \textbf{Sensory} & \textbf{Emotional} & \textbf{Associative} \\
\toprule
\multirow{3}{*}{BERT}    & AST     & 13.68/12.01 & 16.09/13.31& 27.30/23.89 & 0.4952/0.4514   & 13.26/11.42   & 14.21/12.96   & 12.69/11.04       \\
                         & WavVec  & 11.75/10.96 & 15.78/13.16 & 26.58/23.16 & 0.4916/0.4487   & 11.41/10.59  & 13.69/12.01   & 11.46/10.25      \\
                         & EnCodec & 12.64/10.68 & 16.10/13.30 & 27.42/23.75 & 0.4956/0.4574   & 11.48/10.08   & 12.89/11.76   & 11.36/10.07       \\
\midrule
\multirow{3}{*}{T5}      & AST     & \cellcolor{green!30}{16.66*/12.25*} & \cellcolor{green!30}{19.14*/13.19*} & \cellcolor{green!30}{30.62*/24.03*} & \cellcolor{green!30}{0.5579*/0.4610*}   & \cellcolor{green!30}{16.85*/11.51*}   & \cellcolor{green!30}{17.30*}/12.98*   & \cellcolor{green!30}{15.16*/11.20*}       \\
                         & WavVec  & 16.01/10.90 & 18.86/13.27 & 29.48/23.97 & 0.5368/0.4563   & 15.95/11.08  & 16.89/12.67   & 12.09/10.21      \\
                         & EnCodec & 15.96/10.98 & 18.87/13.10 & 29.32/23.85 & 0.5316/0.4563   & 15.03/11.19   & 16.18/12.46   & 12.48/10.32       \\
\midrule
\multirow{3}{*}{Mistral} & AST     & {16.03/12.36}  & {18.58/13.60} & {29.92/23.91}  & {0.5527/0.46002}  & {16.29/11.53}   & 16.89/12.87  & {15.09/11.09}      \\
                         & WavVec  & 15.89/10.68 & 18.42/13.28 & 29.27/23.44 & 0.5484/0.4474   & 15.31/10.47   & 15.93/12.66   & 12.12/10.48      \\
                         & EnCodec & 15.28/10.49 & 17.51/13.11 & 28.68/23.26 & 0.5334/0.4460   & {15.11/10.39}   & 15.63/12.62   & {12.36/9.91}      \\
\midrule
\multirow{3}{*}{Llama}   & AST     & \cellcolor{orange!30}{16.54}*/12.37* & \cellcolor{orange!30}{19.03}*/13.67* & \cellcolor{orange!30}{30.47}*/24.32* & \cellcolor{orange!30}{0.5536}*/0.4731*   & \cellcolor{orange!30}{16.59}*/11.60*  & \cellcolor{orange!30}{17.28}*/13.04*   & \cellcolor{orange!30}{15.04}*/11.28*       \\
                         & WavVec  & 16.14/11.11 & 18.36/13.59 & 29.26/24.30 & 0.5488/0.4689   & 16.08/11.67   & 16.89/13.10   & 12.18/10.29       \\
                         & EnCodec & 15.38/10.72 & 15.62/13.60 & 25.93/24.51 & 0.4880/0.4690    & 16.33/10.85   & 16.40/12.62    & 11.66/10.29   \\
\bottomrule
\end{tabular}}
\caption{
Performance evaluation of framework trained on all categories (sensory, emotional, and associative). The three right columns show P@10 of the framework trained on all three categories but tested on individual categories. For each metric, the number before and after -/- indicate the performance results obtained after filtering out low-agreement data and performance on the full dataset, respectively.
The \colorbox{green!30}{best} and \colorbox{orange!30}{second-best} results are highlighted with green and orange. Significance differences of p < 0.05 using t-test are indicated by *.}
\label{tab:main_result}
\end{table*}
\begin{table*}[t]
\centering
\resizebox{\textwidth}{!}{
\begin{tabular}{cccccccccccccc}
\toprule
                         &         & \multicolumn{4}{c}{\textbf{Sensory}}     & \multicolumn{4}{c}{\textbf{Emotional}}     & \multicolumn{4}{c}{\textbf{Associative}} \\
                         &         & P@10  & R@10  & mAP@10 & nDCG@10 & P@10  & R@10  & mAP@10 & nDCG@10 & P@10  & R@10  & mAP@10 & nDCG@10 \\
\toprule
\multirow{3}{*}{BERT}    & AST     & 15.17/12.65 & 15.86/14.18 & 27.62/24.30 & 0.5116/0.4709   & 15.62/12.92 & 17.80/14.22 & 28.13/25.66 & 0.5138/0.4736 & 14.26/11.89 & 15.20/14.10 & 27.32/24.01 & 0.5022/0.4682   \\
                         & WavVec  & 14.39/11.43 & 15.10/13.89& 26.84/23.75 & 0.5033/0.4601   & 14.49/11.16 & 15.89/ & 27.96/24.15 & 0.5018/0.4605  & 12.11/10.88 & 14.39/13.04 & 26.19/23.26& 0.4928/0.4601   \\
                         & EnCodec & 14.63/11.37 & 15.25/13.61 & 27.10/23.12 & 0.5035/0.4536   & 15.48/11.47 & {17.88/13.48} & 27.16/24.37 & 0.5032/0.4625  & 13.08/10.22& 14.57/ & 26.69/23.68 & 0.4933/0.4644  \\
\midrule \multirow{3}{*}{T5}      & AST  &\cellcolor{green!30} {17.24*/13.48} & \cellcolor{green!30}{17.65*/14.61} & \cellcolor{green!30}{30.17*/25.58} &\cellcolor{green!30} {0.5266*/0.4779}   &\cellcolor{green!30} {18.89*/13.74} &\cellcolor{green!30}{20.79*/15.62} & \cellcolor{green!30}{30.56*/27.10} &\cellcolor{green!30}{0.5274*/0.4886}   & \cellcolor{green!30}{17.88*/12.54} &\cellcolor{green!30} {20.21*/15.06} & \cellcolor{green!30}{29.36*/24.57} &\cellcolor{green!30} {0.5248*/0.4751}   \\
& WavVec  & 16.61/12.87 & 16.34/13.66 & 29.87/24.85 & 0.5078/0.4648   & 17.60/12.20 & 18.99/14.75 & 27.24/26.14 & 0.5016/0.4791   & 15.99/11.40 & 17.12/14.10 & 28.33/23.49 & 0.5087/0.4648   \\
                         & EnCodec & 16.47/12.57& 18.48/13.43 & {29.24}/24.48& 0.5025/0.4611    & 16.28/11.99 & 17.45/14.38 & 26.84/26.04 & 0.4996/0.4725   & 15.58/11.11 & 17.01/13.74& 27.80/23.20& 0.5034/0.4613   \\
\midrule
\multirow{3}{*}{Mistral} & AST     & {16.62/12.87} & \cellcolor{orange!30}{17.64*/14.48} & {29.28/24.37} & 0.5137/0.4748   & 16.86/13.07 & 19.34/15.37 & 28.67/26.04 & 0.5108/0.4743   & 16.83/11.08 & \cellcolor{orange!30}{20.07*/14.09} & 28.55/23.20 & 0.5156/0.4610   \\
 & WavVec  & 16.38/12.36 & 17.16/14.11 & 28.52/24.05 & 0.5073/0.4714   & 16.04/12.84 & 16.29/15.03 & 27.01/25.68  & 0.4982/0.4710   & 15.61/10.46 & 18.14/13.70 & 27.79/23.04 & 0.4940/0.4557  \\
                         & EnCodec & 16.48/12.18 & 17.33/13.89& {28.42/23.78} & {0.5106/0.4709}   & {16.38/12.67} & 17.84/14.78 & 28.25/25.61 & 0.5015/0.4705   & 15.76/10.44 & 19.11/13.76 & 27.37/24.12 & 0.5114/0.4547   \\
\midrule
\multirow{3}{*}{Llama}   & AST     & \cellcolor{orange!30}{17.18*/13.28} & {17.47/14.50} & \cellcolor{orange!30}{29.95*/25.21} & \cellcolor{orange!30}{0.5211*/0.4718}  & \cellcolor{orange!30}{18.48*/13.48} & \cellcolor{orange!30}{20.38*/15.21} & \cellcolor{orange!30}{30.31*/26.85} & \cellcolor{orange!30}{0.5217*/0.4852}   & \cellcolor{orange!30}{17.82*/12.15} & {20.05*/14.97} & \cellcolor{orange!30}{29.12*/24.18} & \cellcolor{orange!30}{0.5226*/0.4709}   \\
                         & WavVec  & 16.29/13.07 & 16.20/13.35 & 26.52/24.14 & 0.5099/0.4687   & 16.52/13.11 & 19.26/14.83& 27.02/26.33 & 0.5101/0.4810   & 16.98/12.02 & 20.01/24.67 & 28.79/23.86 & 0.5118/0.4678   \\
                         & EnCodec & 16.67/13.14 & 16.33/13.38 & 29.21/24.17 & 0.5111/0.4654  & 16.69/13.13 & 19.31/14.67 & 29.27/26.21 & {0.5124/0.4812}   & 17.28/12.04 & 20.83/20.56 & 30.86/24.08 & 0.5140/0.4675   \\
\bottomrule
\end{tabular}}
\caption{Performance evaluation of framework trained on individual category. The numbers before and after the "-/-" denote results on the medium/high agreement subset (i.e., after excluding low-agreement data) and the full dataset, respectively. Significant differences using t-tests are marked by *. }.
\label{tab:results_category}
\end{table*}

\subsection{Setup}
We evaluate the proposed contrastive learning framework from two perspectives: (1) performance on combined categories i.e., the framework is trained using all sensory, emotional, and associative categories, and evaluated on both combined and individual categories, as shown in Section~\ref{sec:combined_result}, and (2) performance on individual categories i.e., the framework is independently trained and evaluated for each category, as discussed in Section~\ref{sec:single_result}. For both perspectives, we report results on the full dataset and after removing the low-agreement subset of the data. Additionally, we report the results of a zero-shot generalization test on our proposed framework in Table \ref{tab:zero_test} in the Appendix. 

\subsection{Evaluation Metrics} 
We use common metrics in Information Retrieval (IR), including Precision@10 (P@10), Recall@10
(R@10),  mean Average Precision at 10 (mAP@10), and normalized Discounted Cumulative Gain at 10 (nDCG@10). 
Precision ensures that the most relevant descriptions appear at the top, while recall is essential for capturing all relevant options. mAP@10 provides a single score that takes both the precision and rank of the retrieved descriptions into account. nDCG@10 gives higher weight to relevant items that appear earlier in the top 10 results and normalizes the weighted score to be between 0 and 1. Together, these comprehensive metrics provide valuable benchmarks for various downstream applications in haptic caption retrieval. 

\subsection{Implementation Details}
All experiments run with one NVIDIA RTX A100 GPU. We perform a light grid search over the hyperparameters, learning rates $\alpha \in \{10^{-3}, 10^{-4}, 10^{-5}\}$, temperature coefficients $\tau \in \{0.07, 0.1\}$, $n=\{1,2,3,4,5\}$, and $m=\{1,2,3,4,5\}$. The model achieves optimal performance when $\alpha=10^{-3}$, $\tau=0.1$, $n=3$ and $m=2$ for best combination of T5 and AST. We set $K=10$, $\kappa=100$, $d_{1}=768$, $d_{2}=768$, and common dimension $d=768$. The batch size is set to 128, and training is conducted for 15 epochs, taking approximately 4 to 12 hours. The optimal hyperparameters with the best epoch for early stopping are determined using a validation set. 

We initialize the parameters of haptic encoder (audio model) and fine-tune last $n$ and $m$ layers of text encoder (LLM) and haptic encoder, respectively (See hyperparameter analysis in Appendix \ref{sec:hyperparameter_analysis}). In our proposed model, the trainable parameters consist of the last $n$ layers of the text encoder and last $m$ layers of haptic encoder, and two linear projection layers. The data is divided into 70\% training, 10\% validation, and 20\% test subsets for the full dataset and the medium/high agreement subset (See Appendix \ref{sec:more_implementation} for more implementation details).

\section{Results}
\subsection{Performance of Model Trained on Combined Categories}
\label{sec:combined_result}
Table \ref{tab:main_result} presents the performance of the proposed framework trained in all categories combined, utilizing separately four text model architectures with three fine-tuned audio models. Also, we report the performance of the model trained on all categories when evaluated separately for the sensory, emotion, and association categories.

First, the model after filtering out low-agreement data performs better, improved by average around 20\% on all metrics, indicating filtering is effective and helpful to performance. Second, our model performs best on the emotional category, followed by performance in sensory category, with the poorest performance in associative category across combinations of language models (e.g., BERT, T5, Mistral, and LLaMA) and AST. We speculate that higher agreement enhances performance, as emotional descriptions generally yield higher agreement, while associative descriptions tend to yield lower agreement. Third, the combination of T5 and AST, as well as Llama and AST, shows superior overall performance across multiple metrics. Specifically, T5 with AST achieves the highest P@10 (16.66), R@10 (19.14), mAP@10 (30.62), and nDCG@10 (0.5579) across all three categories. Llama with AST closely follows in all categories and metrics. In summary, the proposed contrastive learning framework aligns haptic signals and their textual descriptions. For modal encoder, BERT lags behind T5 and Llama in the haptic caption retrieval task, showing certain limitations. 
T5 excels in natural language understanding tasks, while Llama 3.2 performs strongly in natural language generation \cite{fu2023decoder}, leading to T5 slightly outperforming Llama. Regarding haptic encoder, AST is an effective haptic feature extractor and can be paired with LLMs (e.g., T5 and Llama) for the haptic caption task. WavVec is trained on speech corpora, while EnCodec is designed to restore the original signal after compression. 
See the lower performance for training a vibration-specific encoder on HapticCap in Appendix \ref{sec:pretrain}.

\subsection{Performance of Model Trained on Individual Category}
\label{sec:single_result}
We also evaluate the performance of the proposed framework trained in a single category (i.e., sensory, emotional, and associative) across different combinations of pretrained language models and pre-trained audio models. The results are presented in Table \ref{tab:results_category}. Among the models, T5 with AST consistently outperforms others across multiple metrics, particularly excelling in all three categories with the highest scores across the board. In comparison, Llama with AST shows the second-best performance on most metrics. As exceptions, Mistral with AST outperforms Llama with AST in R@10 for the sensory and associative categories. The performance of T5+AST and Llama+AST aligns with the results obtained from training on all categories. We speculate this is because the AST encoder processes spectrograms, capturing frequency-time patterns aligned with human perception of sound and haptics. In contrast, Wav2Vec and EnCodec use raw waveforms, focusing on speech features or compression, and lack global frequency-time representations. Furthermore, we conduct zero-shot generalization experiments and report the results in Appendix \ref{sec:zero-shot_generalization}.


\section*{Conclusion}
In this paper, we introduce the haptic-caption retrieval task to link user descriptions to vibration haptic signals. To support this task, we present a large haptic caption dataset called HapticCap, carefully designed and validated to ensure diversity and quality. We also report results from a supervised contrastive learning framework that incorporates various pre-trained models for text and haptic, aligning user descriptions with vibration signals. 
Our results suggest that haptic-caption retrieval is a challenging task for existing pre-trained models. 
Our experimental results and dataset provide a baseline for future work in this area and pave the way for the development of sensory language models for haptics. 
We are also working on haptic captions with an end-to-end generative framework that takes vibration signal as input and output corresponding captions, aiming to further contribute to the NLP and Haptics communities in our future work.

\section*{Limitations}
First, HapticCap is fully manually annotated through a time-intensive process. While it represents the largest and most diverse collection of vibration signal–caption pairs to date, it may still not encompass the full range of possible signals and user experiences. 
Additionally, user descriptions are only available in English, and data collection is restricted to Denmark and the United States, potentially introducing bias due to geographical limitations. Second, the contrastive learning framework we adopt is a well-established method for aligning multiple modalities and has shown strong performance on various multimodal tasks. As the first work to explore the integration of language and vibration, leveraging this framework offers a solid foundation for cross-modal alignment. Yet, its maturity limits the novelty of our approach.

\section*{Acknowledgement}
We sincerely thank the volunteers for their generous contributions and invaluable efforts in providing high-quality data annotation, which has been instrumental in supporting our research. This work was supported by research grants from VILLUM FONDEN (VIL50296) and the National Science Foundation (\#2339707).



\section*{Ethical Considerations}
The collection and management of personal data in our study were approved by the university's Institutional Review Board (IRB). Additionally, the collection of user information and haptic experience data was conducted with the consent of participants. Participants used pseudonyms, and no identifying information was included in the dataset. We reviewed the haptic descriptions to ensure they do not contain any offensive content.

\bibliography{acl_latex}
\clearpage
\appendix

\section{Additional Experiments}

\subsection{Pretraining a Haptic Encoder}
\label{sec:pretrain}
We also train a haptic-specific encoder from scratch using our vibration signals, with AST and Encodec architectures, respectively. Furthermore, we integrate this encoder with T5 and LLaMA in our contrastive learning framework and evaluate their performance on the haptic caption retrieval task, as shown in Table \ref{tab:pretrained_study}. 

\subsection{More Implementation Details}

\begin{table}[!h]
\centering
\resizebox{\linewidth}{!}{
\begin{tabular}{ccccccc}
\toprule
                         &         & {\bf All Categories} & {\bf Sensory} & {\bf Emotion} & {\bf Association} \\
                         \toprule
&{\bf Ours}&{\bf 16.66} & {\bf 16.85}  & {\bf 17.30}  & {\bf 15.16}     \\
\midrule
&T5+AST& 14.19\textcolor{red}{\(\downarrow\)} &14.68\textcolor{red}{\(\downarrow\)} & 15.12\textcolor{red}{\(\downarrow\)} & 11.53\textcolor{red}{\(\downarrow\)}\\
&Llama+AST& 14.37\textcolor{red}{\(\downarrow\)} &14.74\textcolor{red}{\(\downarrow\)} & 15.26\textcolor{red}{\(\downarrow\)} & 11.61\textcolor{red}{\(\downarrow\)}\\
&T5+Encodec& 12.67\textcolor{red}{\(\downarrow\)} &13.19\textcolor{red}{\(\downarrow\)} & 14.06\textcolor{red}{\(\downarrow\)} & 10.44\textcolor{red}{\(\downarrow\)}\\
&Llama+Encodec& 13.25\textcolor{red}{\(\downarrow\)} &13.43\textcolor{red}{\(\downarrow\)} & 13.71\textcolor{red}{\(\downarrow\)} & 10.26\textcolor{red}{\(\downarrow\)}\\
\bottomrule
\end{tabular}}
\caption{Results on pretraining audio models (e.g., AST) with haptic signals and integrating with large language models (e.g., T5, Llama) on P@10 metric.}
\label{tab:pretrained_study}
\end{table}

Compared to fine-tuning pretrained audio models, the performance of the model with a haptic-specific encoder performance drops notably. This is perhaps because the fine-tuned audio models better learn temporal and rhythmic features, which are shared between vibrations and audio~\cite{bernard2022rhythm}, from much larger audio datasets than HapticCap, and the fine-tuning further adapts them to haptic signals. 

\begin{table*}[!t]
\resizebox{\textwidth}{!}{
\begin{tabular}{cccccccccccccc}
\toprule
                         &         & \multicolumn{4}{c}{\textbf{Sensory}}     & \multicolumn{4}{c}{\textbf{Emotional}}     & \multicolumn{4}{c}{\textbf{Associative}} \\
                         &         & P@10  & R@10  & mAP@10 & nDCG@10 & P@10  & R@10  & mAP@10 & nDCG@10 & P@10  & R@10  & mAP@10 & nDCG@10 \\
\toprule
\multirow{3}{*}{T5+AST}    & Sensory     &  18.16 & 20.23 & 31.32 & 0.5318 &    12.47     &  15.52    & 23.57       &  0.4607    &     11.71        &  13.22    &    20.65  &   0.4461   \\
                           & Emotion     & 11.65        &  13.10    & 21.09      & 0.4467     &  17.45& 20.38 & 29.56 & 0.5298 & 11.01      & 13.08    &  19.25    &   0.4381 \\
                           & Association &     10.10    &   11.89   & 20.56       &  0.4241    &    10.08     &  11.69    &   20.81    &  0.4309    &      17.01 & 20.36 & 29.49 & 0.5225  \\
\midrule
\multirow{3}{*}{Llama+AST} & Sensory     &  18.19 &19.31 & 28.68 & 0.5208   &   12.19      &  15.49    &  23.24      & 0.4593     &      11.68       &   13.14   &  20.36      &   0.4427   \\
                           & Emotion     &   11.24      &  13.05    & 20.88       &  0.4416    &  14.75 & 17.16 & 28.36 & 0.5219   &   11.05         &  13.41    &  18.61      &   0.4312   \\
                           & Association &  10.07       & 11.34     &  20.22      &     0.4420   &  10.10   &  10.86   &   20.08     & 0.4326     & 18.16 & 22.32 & 29.64 & 0.5109 \\
\bottomrule
\end{tabular}}
\caption{The performance of zero-shot generation. Row headers denote the combination of language and haptic encoders and the description category used for model training, and column headers denote the category used for testing the model 
without fine-tuning.}
\label{tab:zero_test}
\end{table*}
\subsection{Zero-Shot Generalization}
\label{sec:zero-shot_generalization}
A zero-shot generalization test assesses a model's ability to handle tasks or data it has never encountered during training. In our work, we evaluate the performance of the proposed framework in zero-shot generalization. Specifically, we test the specific-category model's performance on unseen categories. For instance, we train the framework with the sensory description and test it on the emotional and associative categories. The results are presented in Table \ref{tab:zero_test}.

First, models perform best when tested on the same category they were trained on (diagonal values). When tested on other categories, performance drops, indicating a domain shift in description styles. For example, Sensory-trained models struggle with emotional and associative captions. Second, Sensory-trained models perform better across different categories compared to emotion-trained and associative-trained models. Sensory-related descriptions can overlap with both emotional (e.g., ``intense'' can convey strong emotions) and associative (e.g., ``beat'' or ``pulse'' are related to heartbeat). In contrast, emotional and associative categories are more domain-specific, making cross-category adaptation more challenging. These results suggest that feature alignment methods, such as adversarial domain adaptation, could help mitigate the gap between different description styles. In terms of generalization, sensory training provides better generalization than emotional or associative training.

\subsection{Hyperparameter Analysis}
\label{sec:hyperparameter_analysis}
We also conduct experiments with different values for the number of fine-tuning last layer $n$ of text encoder T5. The model achieves the optimal result when $n$=3, both for combined and separate categories. The detailed P@10 results for $n \in \{1,2,3,4,5\}$ are reported in Table \ref{tab:hyperparameter_analysis}. 

\begin{table}[t]
\centering
\resizebox{\linewidth}{!}{
\begin{tabular}{ccccccc}
\toprule
                         &         & {\bf All Categories} & {\bf Sensory} & {\bf Emotion} & {\bf Association} \\
                         \toprule
&n=1& 11.47\textcolor{red}{\(\downarrow\)} &10.78\textcolor{red}{\(\downarrow\)} & 11.91\textcolor{red}{\(\downarrow\)} & 10.36\textcolor{red}{\(\downarrow\)}\\
&n=2& 15.61\textcolor{red}{\(\downarrow\)} &15.36\textcolor{red}{\(\downarrow\)} & 16.82\textcolor{red}{\(\downarrow\)} & 15.19\textcolor{red}{\(\downarrow\)}\\
&n=3& 16.74\textcolor{red}{\(\downarrow\)} &16.59\textcolor{red}{\(\downarrow\)} & 17.28\textcolor{red}{\(\downarrow\)} & 15.44\textcolor{red}{\(\downarrow\)}\\
&n=4& 14.94\textcolor{red}{\(\downarrow\)} &14.51\textcolor{red}{\(\downarrow\)} & 15.64\textcolor{red}{\(\downarrow\)} & 14.25\textcolor{red}{\(\downarrow\)}\\
&n=5& 14.39\textcolor{red}{\(\downarrow\)} &14.16\textcolor{red}{\(\downarrow\)} & 14.86\textcolor{red}{\(\downarrow\)} & 14.11\textcolor{red}{\(\downarrow\)}\\
\bottomrule
\end{tabular}}
\caption{Hyperparameter analysis on fine-tuning layer on P@10 metric.}
\label{tab:hyperparameter_analysis}
\end{table}

\subsection{More Implementation Details}
\label{sec:more_implementation}
We present the hyperparameter setting for the combinations of text encoder (e.g., LLMs T5 and Llama) and haptic encoder (e.g., AST, WavVec and Encodec) as Table \ref{tab:value_hyper}.
\begin{table}[!h]
\centering
\resizebox{0.8\linewidth}{!}{
\begin{tabular}{ccccc}
\toprule
& {n} & {m} & {$\alpha$} & {$\tau$}\\
\toprule
T5 + AST & 3 & 2 & $10^{-3}$ & 0.1\\
T5 + WavVec & 3 & 3 & $10^{-3}$ & 0.1\\
T5 + Encodec & 3 & 3 & $10^{-3}$ & 0.1\\
\midrule
Llama + AST & 2 & 2 & $10^{-3}$ & 0.1 \\
Llama + WavVec & 2 & 3 & $10^{-3}$ & 0.1 \\
Llama + Encodec & 2 & 3 & $10^{-3}$ & 0.1 \\
\bottomrule
\end{tabular}}
\caption{The values of hyperparameters.}
\label{tab:value_hyper}
\end{table}

\section{Dataset Analysis}
\label{sec:datasetanalysis}

\subsection{Diversity Analysis}
\label{sec:diversity}

\begin{figure}[!ht]
\centering
\subfigure{\label{fig:subfig:a}}\addtocounter{subfigure}{-1}
\subfigure[]
{\includegraphics[width=0.46\linewidth]{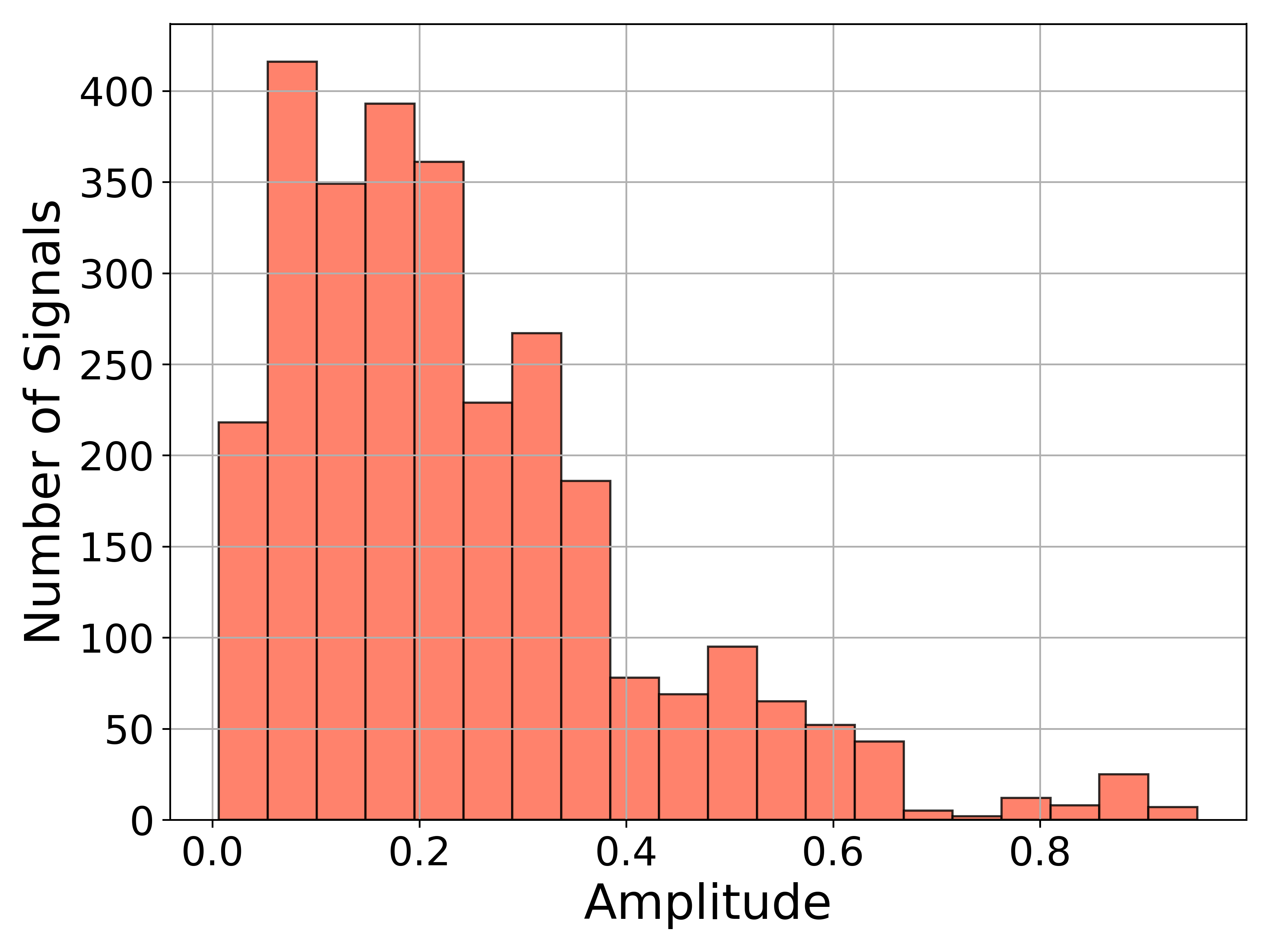}}
\subfigure{\label{fig:subfig:b}}\addtocounter{subfigure}{-1}
\subfigure[]
{\includegraphics[width=0.46\linewidth]{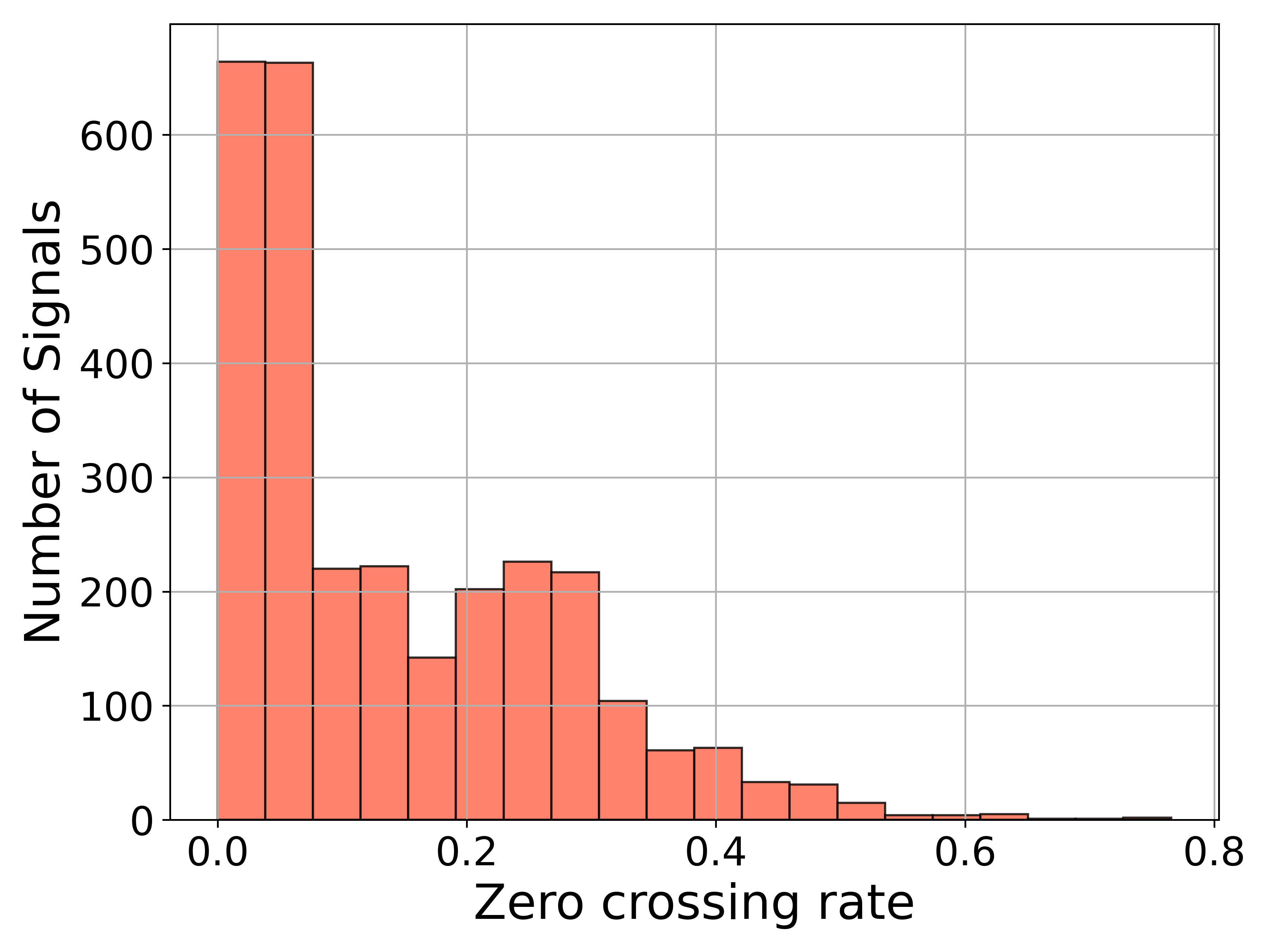}}
\caption{Haptic signal statistics : (a) signal count on amplitude feature, (b) signal count on zero crossing rate.}
\label{fig:haptic_feature}
\end{figure}

\begin{figure}[t]
\centering
\subfigure{\label{fig:subfig:a}}\addtocounter{subfigure}{-1}
\subfigure[sensory]
{\includegraphics[width=0.9\linewidth]{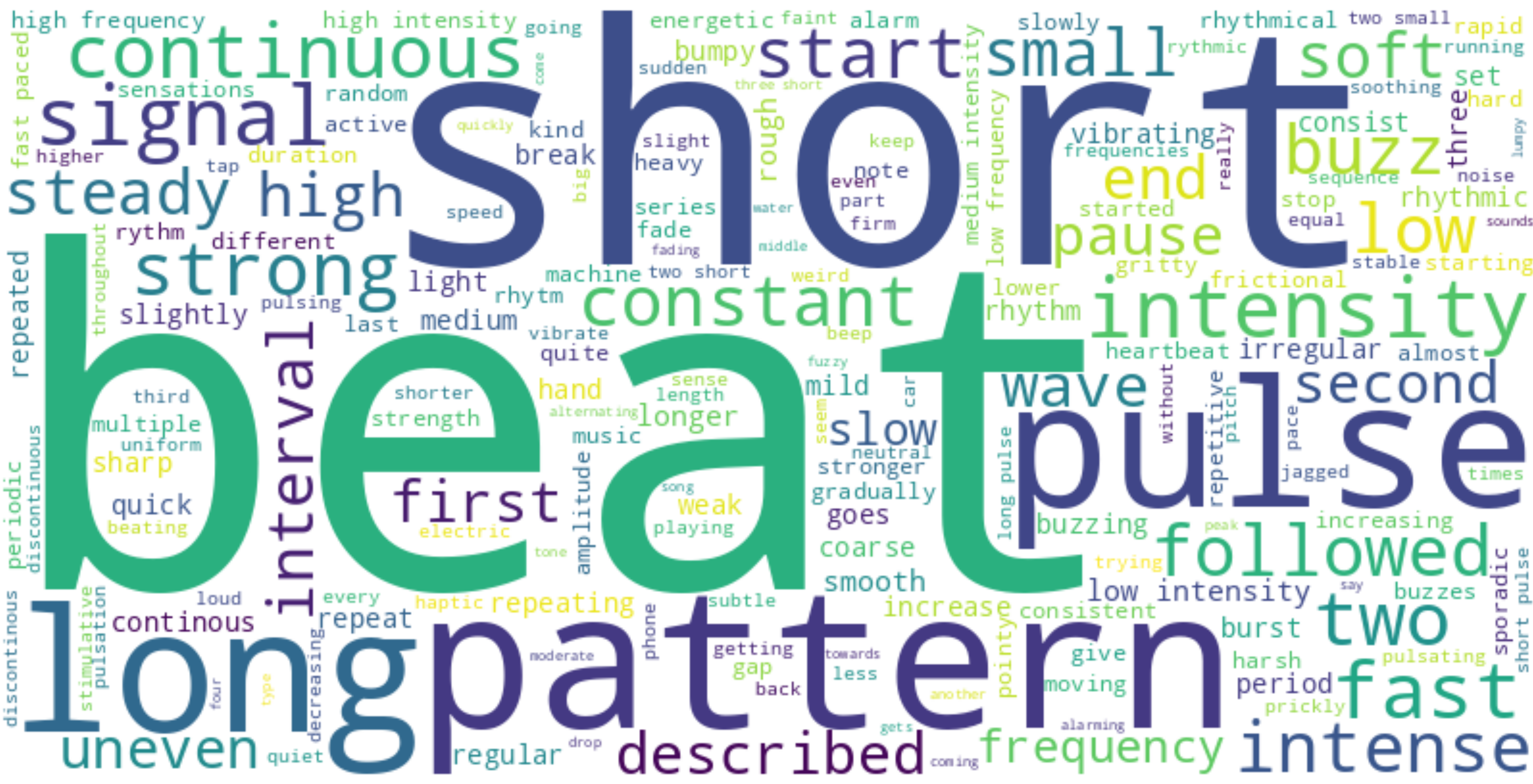}}
\subfigure{\label{fig:subfig:b}}\addtocounter{subfigure}{-1}
\subfigure[emotional]
{\includegraphics[width=0.9\linewidth]{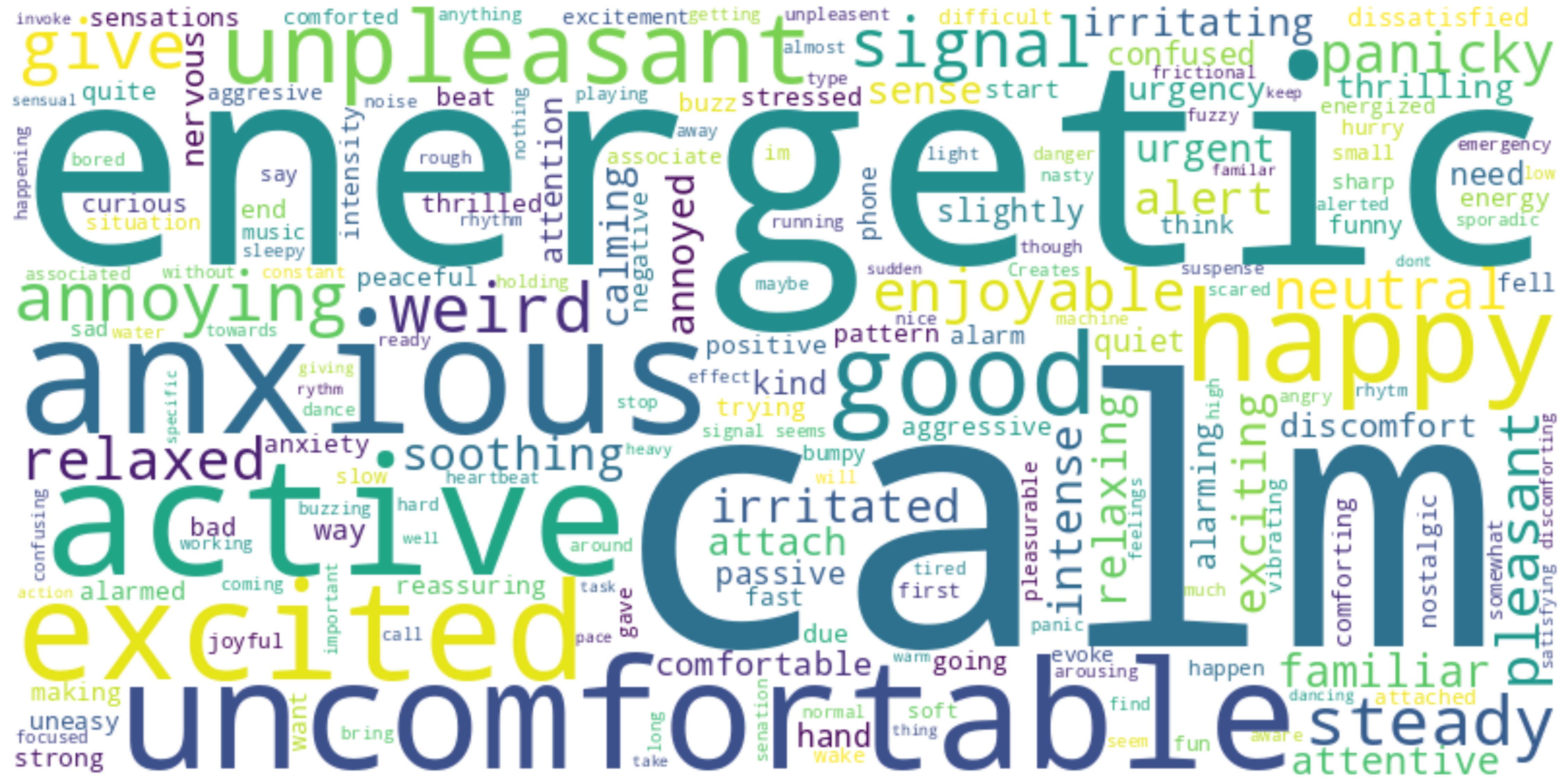}}
\subfigure{\label{fig:subfig:b}}\addtocounter{subfigure}{-1}
\subfigure[associative.]
{\includegraphics[width=0.9\linewidth]{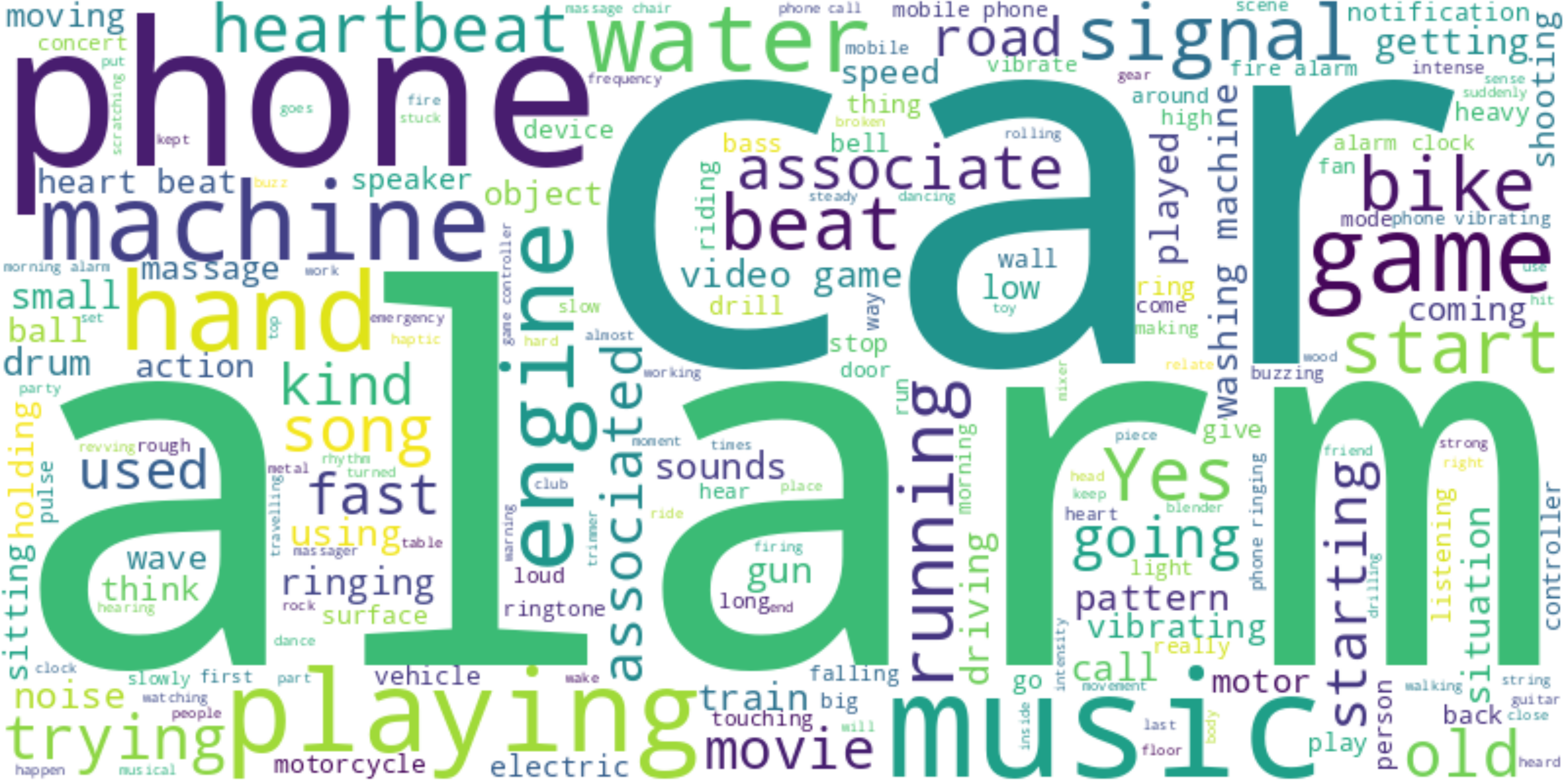}}
\caption{Visualization of word distribution across sensory, emotion, and association aspects.}
\label{fig:word_cluster}
\end{figure}

\begin{figure}[!ht]
\centering
\subfigure{\label{fig:subfig:a}}\addtocounter{subfigure}{-1}
\subfigure[]
{\includegraphics[width=0.46\linewidth]{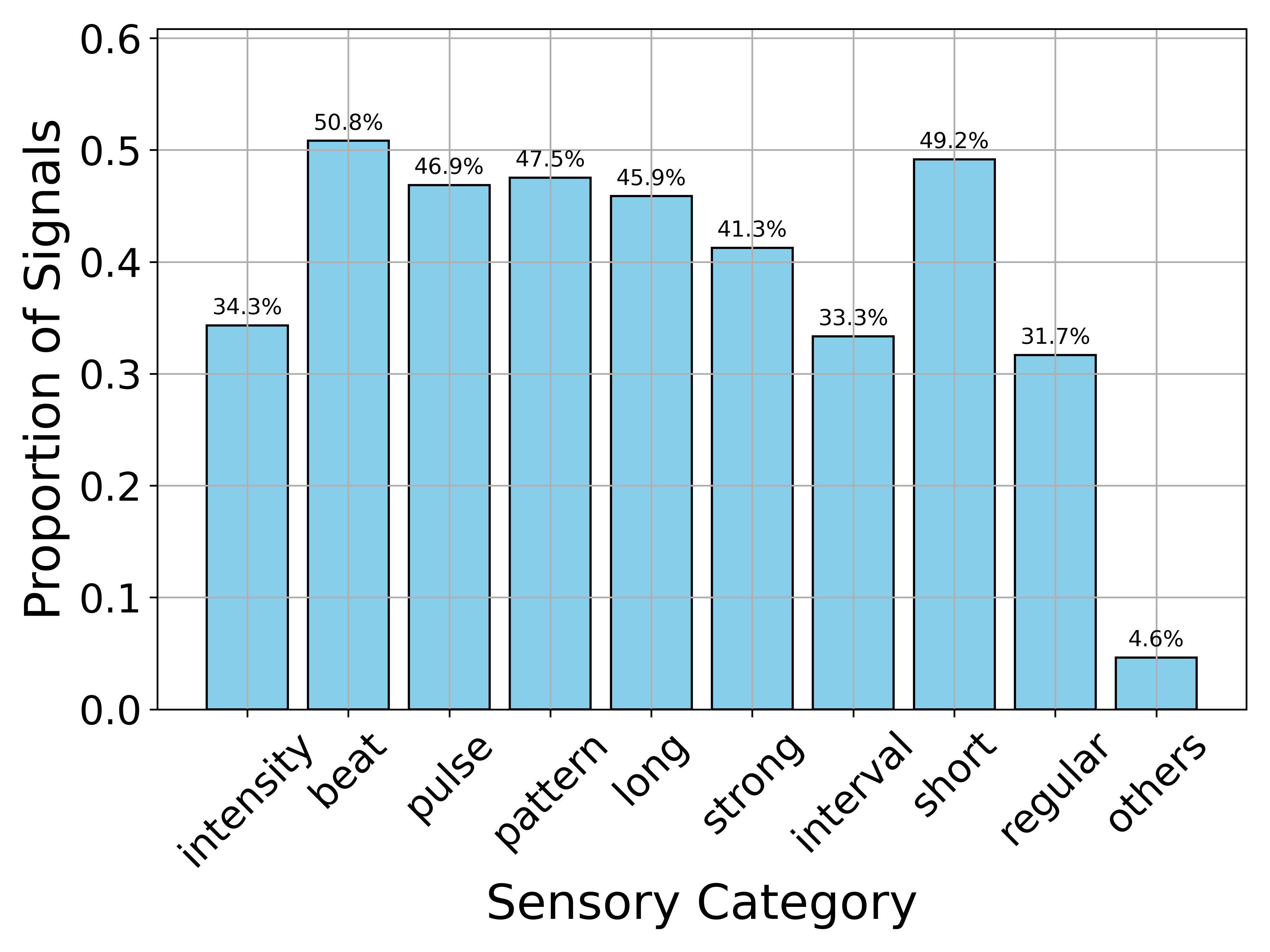}}
\subfigure{\label{fig:subfig:b}}\addtocounter{subfigure}{-1}
\subfigure[]
{\includegraphics[width=0.46\linewidth]{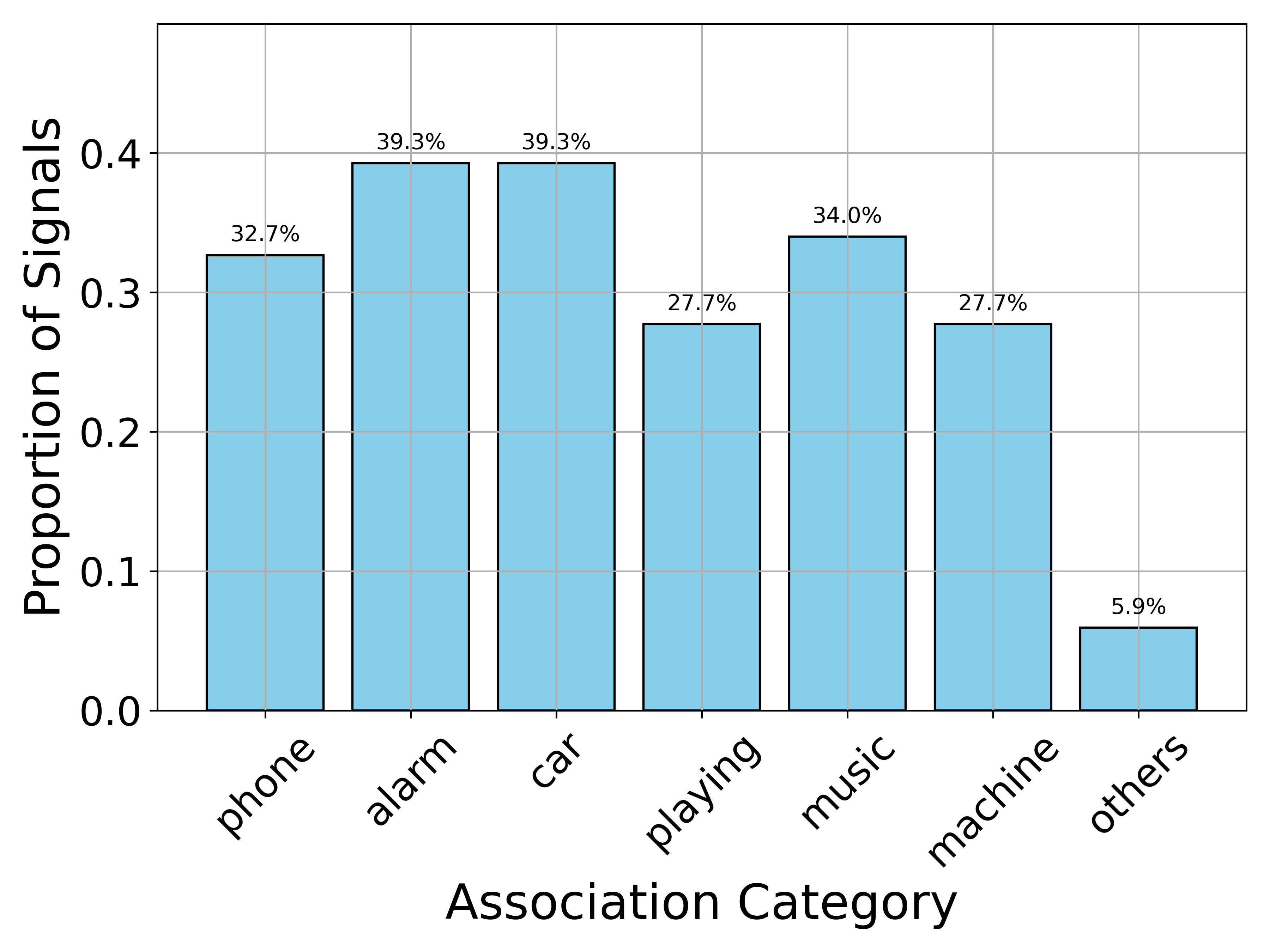}}
\caption{Distribution of vibration signals in HapticCap over (a) sensory words, and (b) associative words.}
\label{fig:sen_ass_distribution}
\end{figure}

We perform a comprehensive diversity analysis of the proposed HapticCap dataset. Figure \ref{fig:haptic_feature} presents the distribution of vibration haptic signal based on amplitude and zero crossing rate features. Figure \ref{fig:word_cluster} shows word-cloud visualizations corresponding to the sensory, emotional, and associative captions. Additionally, Figure \ref{fig:sen_ass_distribution} illustrates the distribution of haptic signals associated with sensory and associative words, which are extracted from their respective word clouds.

Regarding sensory aspect, words such as short, long, strong, beat, pulse, fast, and interval appear prominently, indicating that sensory descriptions focus on measurable and temporal characteristics of haptic signals. These terms suggest that users describe touch stimuli based on rhythm,  intensity, and speed. 
For emotional category, the most dominant words include calm, energetic, anxious, uncomfortable, excited, good, and happy, demonstrating that haptic signals evoke clear emotional responses. The presence of both positive (e.g., calm, energetic) and negative (e.g., anxious, uncomfortable) emotions indicates a diverse range of affective reactions. In term of associative, words like alarm, car, phone, music, game, machine, and heartbeat dominate this category, reflecting the tendency of users to associate haptic sensations with real-world experiences, objects, or events. These associations suggest that haptic signals can be linked to familiar contexts, making them more interpretable. 

Overall, the word clouds highlight the diversity of user experiences and the distinct focus of each aspect: sensory descriptions emphasize physical properties, emotional descriptions capture feelings, and associative descriptions link haptic experiences to real-world references. 

The distribution of signals over sensory and associative words in Figure~\ref{fig:sen_ass_distribution} further suggest the vibration signals in HapticCap can create diverse experiences (see Figure \ref{fig:statistics}a for the emotional category). 
Finally, we calculate distinct-n and n-gram of description to measure the diversity of textual description, shown in Table \ref{tab:desc_diverse}.

\begin{table}[!ht]
\resizebox{\linewidth}{!}{\begin{tabular}{l|c|c|c}
\toprule
                      & \bf Distinct-1/1-Gram\(\uparrow\) & \bf Distinct-2/2-Gram\(\uparrow\) & \bf Distinct-3/3-Gram\(\uparrow\) \\
\toprule
Sensory & 0.0599/7.7289 & 0.3528/12.0394 & 0.6793/13.9565 \\
Emotion & 0.0639/6.9261 & 0.2995/10.2775 & 0.5303/11.8154\\
Association & 0.0868/7.9509 & 0.4065/11.8475 & 0.6794/13.5471 \\
\bottomrule
\end{tabular}}
\caption{Metrics on distinct-n and n-gram of description.}
\label{tab:desc_diverse}
\end{table}

\section{Data Collection and Compensation}
\label{sec:annotation}
\begin{figure}[!h]
\centering
\subfigure{\label{fig:subfig:a}}\addtocounter{subfigure}{-1}
\subfigure[]
{\includegraphics[width=0.49\linewidth]{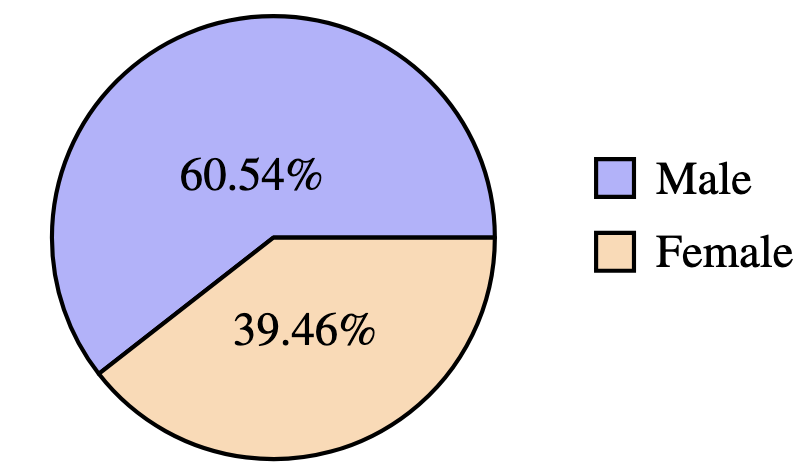}}
\subfigure{\label{fig:subfig:b}}\addtocounter{subfigure}{-1}
\subfigure[]
{\includegraphics[width=0.46\linewidth]{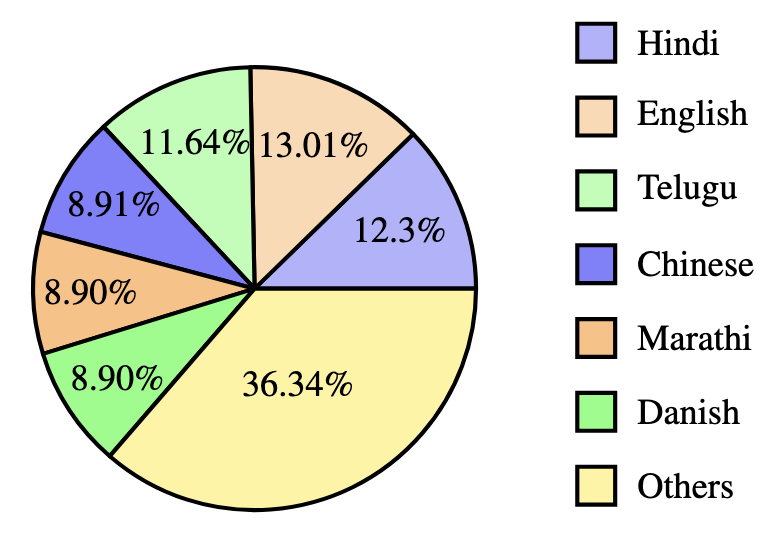}}
\caption{Distribution of (a) gender and (b) native language of users collected in the HapticCap dataset. All users were fluent English speakers.}
\label{fig:background}
\end{figure}

In this work, the annotators come from diverse backgrounds, including students, haptic designers, and researchers. Figure \ref{fig:background} visualizes participant backgrounds.  Each participant described 16 signals (one group) within one hour and received \$15 USD in cash or an equivalent value in gifts (e.g., chocolate) as compensation for their time. This compensation rate is above the minimum local wage. The data collection period spans approximately 11 months, including the user study and the creation of new haptic signals.

\begin{figure}[H]
\centerline{\includegraphics[trim={0 0 1.9cm 0},clip,width=0.5\textwidth]{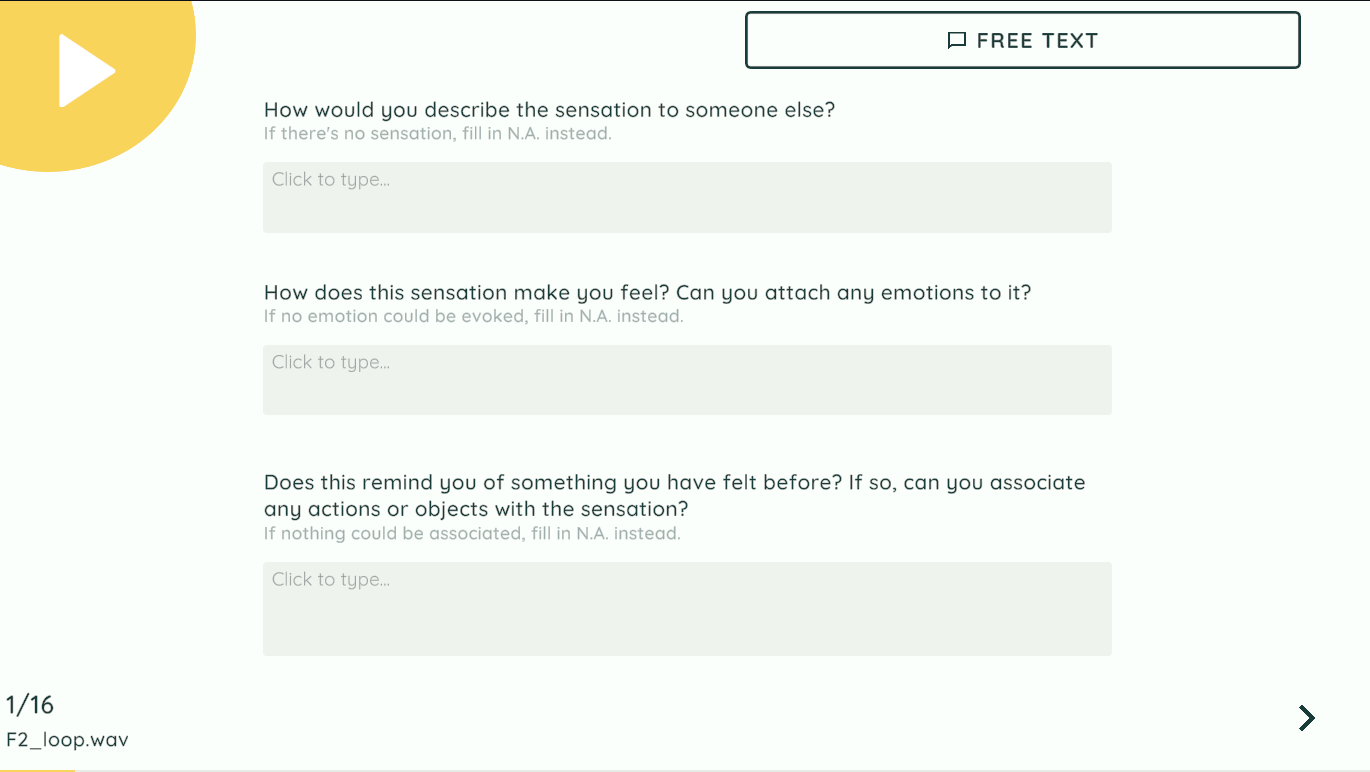}}
\caption{Illustration of the data collection interface.}
\label{fig:collection_platform}
\end{figure}

To ensure high-quality annotations, we collected this dataset by having users feel vibrations on VR handheld devices in a controlled lab setting. Figure~\ref{fig:collection_platform} shows the graphical user interface that participants used to play the haptic signals and includes the questions and instructions to write sensory, emotional, and associative descriptions. Table~\ref{tab:case_illustration} presents example sensory, emotional, and associative descriptions from three different participants in our dataset.

We divide all haptic signals into 19 groups/sets, with each set containing 16 signals. The first two sets have a larger number of participants as we collected more descriptions for these two sets of haptic signals when we started the data collection. Based on our initial analysis of these descriptions, we found that descriptions from 10 users per signal provide enough variation, and thus we continued with describing each signal by 10 users for the following sets. We include the additional descriptions for the first two sets in HapticCap to enable future work to do further analysis of the individual differences and subjectivity in haptic descriptions. 

\begin{table}[H]
\centering
\footnotesize
\resizebox{0.5\textwidth}{!}{
\begin{tabular}{lll}
\toprule
& \textbf{Category} & \textbf{Description}\\
\midrule
\multirow{3}{*}{$P_{1}$} & sensory     & Its frequency is not very strong but very regular. \\
                    & emotion     & I feel little bit impatient.                                                      \\
                    & association & It reminds me of my heartbeat after an intense workout.                            \\
\midrule
\multirow{3}{*}{$P_{2}$} & sensory     & feeling is like beating of a heart that is beating fast.                           \\
                    & emotion     & makes me feel a bit stressed , like I need to hurry.                               \\
                    & association & It is like the heartbeat after running fast.                              \\
\midrule
\multirow{3}{*}{$P_{3}$} & sensory     & The sensation was short and with a light vibration.                           \\
                    & emotion     & I feel being made hurring up by the vibrations.                              \\
                    & association & It reminds me of the feeling of sitting on a motor boat.                             \\
\bottomrule
\end{tabular}}
\caption{Example haptic descriptions in sensory, emotional, and associative categories. $P_{i}$ denotes participant ID.}
\label{tab:case_illustration}
\end{table}

\end{document}